# Enhancing Robustness in Robot-Environment Interactions through Passive Compliant Degrees of Freedom: A Hybrid Position-Force Control Approach with Feedback Linearization

1st Rahman Ardakanian
*Department of Mechanical Engineering,* Ferdowsi University of Mashhad, Mashhad, Iran
ardakanian.rahman@mail.um.ac.ir

2nd Iman Kardan,
Center of Advanced Rehabilitation and Robotics Research, Department of Mechanical Engineering, Ferdowsi University of Mashhad
Mashhad, Iran
iman.kardan@um.ac.ir

3rd AliAkbar Akbari
*Department of Mechanical Engineering,* Ferdowsi University of Mashhad, Mashhad, Iran
akbari@um.ac.ir

4th Ali Mousavi
Department of Computer Engineering, Ne. C, Islamic Azad University, Neyshabur, Iran
mousavi@iau.ac.ir

***Abstract* - Robot-environment interactions in dynamic or unstructured settings are often degraded by impact shocks, vibrations, and uncertainties in contact geometry and mechanical properties. This paper proposes an interaction architecture that combines feedback-linearized hybrid position-force control with a passive compliant degree of freedom embedded at the end-effector. Unlike conventional hybrid position-force control, which relies mainly on active feedback, force sensing, and gain tuning, the proposed architecture uses a physical spring-damper interface to store and dissipate impact energy at the contact point before high-frequency shocks propagate to the actuated joints and force-control loop. The approach is evaluated in MATLAB/Simulink on a 2-DOF planar manipulator with three end-effector configurations: rigid, spring-only, and spring-damper. Results under fixed and time-varying interaction conditions show that the spring-damper configuration provides stronger attenuation of contact-induced oscillations, lower force and velocity error variance, and smoother joint-torque response. Representative reductions include 36.5% in fixed-environment tangential force-error standard deviation, 25.4% in variable-environment normal force-error standard deviation, and 41.1% in variable-environment normal velocity-error standard deviation.**

## I. INTRODUCTION

Robot-environment interaction lies at the core of modern robotic applications such as assembly, polishing, surface finishing, surgery, and human-robot collaboration. In these tasks, the robot must not only follow precise trajectories but also regulate the forces it applies to its surroundings. However, uncertainties in mechanical properties and surface geometry make this interaction inherently difficult. Even small mismatches between expected and actual contact conditions can lead to inaccurate force regulation or severe impact shocks, raising concerns about safety, performance, and long-term durability of the robot and its environment.

Classical position control performs exceptionally well in free space, offering accurate trajectory tracking and simplicity in implementation. Yet, once the manipulator encounters constraints, its inability to regulate contact forces may result in large impacts or unstable behaviors. Conversely, force control focuses on regulating interaction forces through sensory feedback, but it offers no guarantee of maintaining the desired position along unconstrained directions. This fundamental trade-off motivated the development of hybrid position-force control, pioneered by Raibert and Craig [3], which allocates motion control to unconstrained task directions and force control to constrained ones. While this hybrid paradigm has become a standard for interacting robots, it still suffers from practical shortcomings when exposed to uncertainties or abrupt impacts.

A parallel and influential research line emphasizes passivity-based control, originally formalized by Ortega and Spong [1], which frames stability in terms of the system's energy exchange with the environment and successfully applied to force-motion regulation of robot [2]. The central idea is to prevent the robot from generating excess energy that could destabilize the interaction. However, even passivity-based schemes may show degraded performance under strong impacts, rapid changes in environmental stiffness, or contact-loss events. These issues are aggravated when static controller gains, usually tuned for a specific working state, do not generalize to diverse working conditions.

To address such limitations, researchers have also explored virtual compliance through impedance and admittance control. Hogan's seminal impedance control framework [4,11] models the interaction as a virtual mass-spring-damper element, enabling predictable force-motion behavior without requiring full environmental knowledge. Despite its practical value, impedance-based approaches rely heavily on sensing accuracy and may struggle during high-impact collisions, where virtual compliance can provide limited bandwidth and may fail in absorbing the shock energy. Related dynamic compensation strategies, such as feedback linearization (e.g., Slotine and Li [5]), improve trajectory tracking under nonlinear robot dynamics but still assume relatively smooth interactions, making them susceptible to modeling uncertainties and abrupt forces.

Recent studies have further extended compliant force control and passivity-aware interaction control toward uncertain and human-interactive environments. Variable admittance control has been used to adapt robot compliance when the human or environmental stiffness changes during interaction, and recent work has specifically addressed oscillation suppression, human intention adaptation, and experimental validation in physical human–robot interaction [21]. Safety-critical admittance frameworks have also been proposed by combining admittance control with control barrier functions or optimization-based constraints to maintain compliant behavior without violating safety limits [22]. In addition, robust variable admittance control has been investigated for human–robot co-manipulation under unknown payloads and uncertain interaction conditions [23], while sliding-mode and passivity-related variable admittance strategies have been proposed to improve robustness and stability in physical interaction [24]. Recent surveys also confirm that compliant force control remains a central approach for safe and accurate robot interaction with dynamic and uncertain environments [25].

Recent studies published in the Journal of Applied and Computational Sciences in Mechanics also address related aspects of robot modeling, nonlinear control, contact interaction, and passive/mechanical design. Aalipour et al. used feedback linearization for nonlinear motion control of a spherical robot under parameter uncertainty and disturbance torque [26]. Ebrahimi et al. developed a constrained force-control framework for an aerial manipulator interacting with an environment of unknown stiffness [27]. Yousefzadeh and Shafei modeled contact and friction forces in flexible robotic arms, emphasizing the role of viscoelastic collision forces, friction, and vibration during robot-environment interaction [28]. Varedi-Koulaei et al. showed that joint clearance can introduce impact, shock, vibration, and significant changes in velocity and acceleration in planar parallel manipulators [29]. Bamdad and Mardani also demonstrated how a passive joint can be used in a mobile robot structure to provide mechanical functionality without an additional actuator [30]. These studies confirm the relevance of nonlinear robot control, contact-force modeling, vibration effects, and passive mechanical structures within the journal scope; however, they do not directly compare rigid, spring-only, and spring-damper passive end-effector interfaces under the same feedback-linearized hybrid position-force controller, which is the focus of the present work.

Despite these advances, including the above studies on nonlinear control, force control, contact modeling, and passive structural design, most recent works still focus on virtual compliance, adaptive admittance parameters, optimization-based safety constraints, or software-level passivity enforcement. In contrast, the present work investigates the complementary role of physical passive compliance at the contact interface. By embedding a passive spring-damper DOF directly at the end-effector, the proposed architecture provides mechanical energy storage and dissipation before the contact disturbance is fully transmitted to the active hybrid position-force controller. This distinction motivates the comparative analysis of rigid, spring-only, and spring-damper end-effector interfaces under the same feedback-linearized hybrid control structure.

The modeling of environment behavior has similarly evolved from simple elastic approximations [6] to viscoelastic models such as Maxwell [7], Kelvin-Voigt [8], Kelvin-Boltzmann [9], and nonlinear Hunt-Crossley contacts [10]. While these models enhance simulation fidelity, they also highlight how widely environmental characteristics can vary, suggesting that control strategies relying solely on active algorithms may remain brittle in unpredictable settings. Since the main objective of the proposed architecture is to improve robustness in uncertain contact, evaluating the response under changes in environmental stiffness and damping is necessary to determine whether the passive compliant interface remains effective beyond a single nominal environment.

Recent studies in collaborative and industrial scenarios further show that active and virtual compliance methods often require accurate sensing, online parameter adaptation, stability monitoring, or optimization-based constraints to maintain safe interaction under changing contact conditions [14], [15], [21]–[24]. However, these approaches mainly address compliance through software-based adaptation, whereas systematic comparison of physical end-effector compliance configurations under the same hybrid force-position controller remains limited. A notable gap in the literature is the limited exploration of physical passive degrees of freedom (DOF) as part of the robot's hardware interface. Most works rely exclusively on virtual compliance, despite physical passive elements such as springs or dampers being naturally effective at absorbing shocks, increasing safety, and reducing sensitivity to uncertainties. Moreover, very few studies have systematically compared how different passive configurations influence interaction performance across environments with variable stiffness or damping.

These observations motivate a new approach that integrates the strengths of both physical and algorithmic solutions. By incorporating a passive compliant DOF into the end-effector and combining it with feedback-linearized hybrid position control, a robot can benefit from robust tracking in free space, stable force regulation in contact, and inherent mechanical shock absorption during impacts. Such a hybrid physical-algorithmic framework has the potential to mitigate overshoot, reduce settling time, handle diversified mechanical uncertainties, and maintain safe interactions even during abrupt or unmodeled collisions.

The present work aims to fill this gap by investigating the role of passive DOF configurations with spring-only and spring-damper settings, within a feedback-linearized hybrid position-force control scheme. To do this, a passive compliant DOF is added to the end-effector of a 2-DOF planar robot and its performance is compared for three different configurations of: (1) rigid end-effector (no passive DOF), (2) spring-only passive DOF, and (3) spring-damper passive DOF. The analysis shows that embedding a mechanical passive compliant DOF can improve simulated interaction behavior by reducing contact-induced oscillations and error variance.

The main novelty of this work is not the classical hybrid position-force control law itself, but the proposed integration of a physical passive compliant DOF as an active part of the interaction architecture. In conventional hybrid position-force control, contact transients are mainly handled through feedback regulation, force sensing, and controller gain tuning. In the proposed approach, the contact interface itself contributes to robustness by mechanically storing and dissipating impact energy before it excites the actuated robot dynamics. Therefore, passive compliance is treated as a physical energy-management mechanism integrated with the hybrid controller, rather than as a separate protective attachment or a purely virtual impedance behavior.

From an implementation perspective, the proposed method is intended to be realizable using standard robotic hardware equipped with joint encoders, a force/torque sensor at the end-effector, and a passive spring-damper interface mounted along the contact direction.

The specific contributions of this paper are summarized as follows.

First, a hybrid physical-control interaction architecture is proposed by integrating a passive compliant end-effector DOF with feedback-linearized hybrid position-force control.

Second, the passive DOF is used as a mechanical energy-management element at the contact interface, allowing impact energy to be stored and dissipated before entering the active control loop.

Third, rigid, spring-only, and spring-damper end-effector configurations are evaluated under the same control architecture in both fixed and variable environment conditions, allowing the effects of elastic energy storage and viscous energy dissipation to be distinguished.

Fourth, the comparative results show that the spring-damper interface reduces contact-induced oscillations, force/velocity error variance, and joint-torque fluctuations relative to the rigid and spring-only cases in both fixed and time-varying interaction scenarios.

The paper is structured as follows: Section II describes system modeling, trajectories, environment, and control overview. Section III details methodology with hybrid control and feedback linearization. Section IV outlines simulation setup. Section V presents results, analysis, limitations, and implications. Section VI concludes the paper with findings, future work, and broader impact.

## II. System Description and Modeling

This section provides foundational modeling and concepts derived from the system dynamics.

### A. Robot Dynamics

We consider a 2-DOF planar manipulator. The 2-DOF planar manipulator is selected as a benchmark model because it captures the essential coupling between joint-space dynamics, task-space motion, and normal/tangential contact interaction while remaining simple enough for transparent analysis of the passive compliant interface. The selected link lengths and masses, reported in Appendix Table I, represent a small laboratory-scale planar manipulator and are within the range commonly used in simulation studies of robotic contact control. The system dynamics are derived using the Lagrange method [18], incorporating the passive elements. The governing equations of motion of a manipulator with $n$ joints in joint-space can be represented as,

$$M(\mathbf{q})\ddot{\mathbf{q}} + C(\mathbf{q},\dot{\mathbf{q}})\dot{\mathbf{q}} + \mathbf{G}(\mathbf{q}) + \boldsymbol{\tau}_f(\dot{\mathbf{q}}) + \boldsymbol{\tau}_d = \boldsymbol{\tau} - J^T\mathbf{F}_{ext}, \qquad (1)$$

where $\mathbf{q} \in \mathbb{R}^{n\times 1}$ is the vector of $n$ joints positions and the coordinates of joint-space, $M(\mathbf{q}) \in \mathbb{R}^{n\times n}$ is the inertia matrix, $C(\mathbf{q},\dot{\mathbf{q}}) \in \mathbb{R}^{n\times n}$ is the Coriolis and centrifugal matrix, $\mathbf{G}(\mathbf{q}) \in \mathbb{R}^{n\times 1}$ is the gravity vector, $\boldsymbol{\tau}_f(\dot{\mathbf{q}}) \in \mathbb{R}^{n\times 1}$ is the vector of Coulomb and Viscous friction torques, $\boldsymbol{\tau}_d \in \mathbb{R}^{n\times 1}$ is the vector of external disturbance torques, $\boldsymbol{\tau} \in \mathbb{R}^{n\times 1}$ is the vector of commanded actuator torques applied at the robot joints. It represents the control input generated by the proposed feedback-linearized hybrid position-force controller and delivered by the joint motors. $J^T\mathbf{F}_{ext}$ is the vector of mapped torques from the external forces applied to the end-effector.

Therefore, $\boldsymbol{\tau}$ denotes the active control effort supplied by the robot actuators, whereas $\boldsymbol{\tau}_f$ , $\boldsymbol{\tau}_d$, and $J^T\mathbf{F}_{ext}$ represent friction, external disturbances, and contact-force effects acting on the manipulator dynamics.

The corresponding end-effector dynamics of a manipulator in the cartesian space can be written as

$$M_O\ddot{\mathbf{p}} + C_O\dot{\mathbf{p}} + \mathbf{G}_O + \mathbf{F}_f + \mathbf{F}_d = \mathbf{F} - \mathbf{F}_{ext}, \qquad (2)$$

where $\mathbf{p} \in \mathbb{R}^{m\times 1}$ is the vector of $m$ coordinates of task-space. The transformed terms are as follows.

- $\mathbf{p} = \mathbf{f}(\mathbf{q})$
- $\dot{\mathbf{p}} = \partial\mathbf{f}(\mathbf{q})/\partial t = \partial\mathbf{f}/\partial\mathbf{q} \times \partial\mathbf{q}/\partial t = J\dot{\mathbf{q}}$
- $\ddot{\mathbf{p}} = \dot{J}\dot{\mathbf{q}} + J\ddot{\mathbf{q}}$
- $M_O = {J^{-1}}^T M J^{-1}$
- $C_O = {J^{-1}}^T (C - MJ^{-1}\dot{J})J^{-1}$ (3)
- $\mathbf{G}_O = {J^{-1}}^T\mathbf{G}$
- $\boldsymbol{\tau}_f = J^T\mathbf{F}_f$
- $\boldsymbol{\tau}_d = J^T\mathbf{F}_d$
- $\boldsymbol{\tau} = J^T\mathbf{F}$

When the Jacobian is non-square or when task-space mapping requires a generalized inverse, the Moore-Penrose pseudo-inverse is used. However, the feedback-linearization formulation assumes that the manipulator operates away from kinematic singularities so that the Jacobian or its pseudo-inverse remains well-conditioned.

- $J^\dagger = (J^TJ)^{-1}J^T$
- $M_O = {J^\dagger}^T M J^\dagger$
- $C_O = {J^\dagger}^T (C - MJ^\dagger\dot{J})J^\dagger$ (4)
- $\mathbf{G}_O = {J^\dagger}^T\mathbf{G}$

### B. End-Effector Configuration

In this paper, three different configurations are analyzed for the end-effector: rigid end-effector with *no* extra passive compliant DOF, end-effector augmented by a spring-only passive compliant DOF, and end-effector augmented by a spring-damper passive compliant DOF.

The passive compliant DOF is assumed to be a translational element mounted along the normal contact direction between the robot wrist and the end-effector tool. The spring is modeled as a linear elastic element, and the damper is modeled as a linear viscous element. The mass and internal inertia of the passive module are neglected compared with the robot links, and the spring-damper deformation is assumed to remain within its linear operating range. Under these assumptions, the passive element contributes contact-dependent elastic and dissipative forces without adding active actuation.

In practical industrial robots, the passive compliant DOF can be implemented as an intermediate mechanical module between the robot flange and the end-effector tool. However, its realization is subject to several design constraints, including allowable tool deflection, payload capacity, added mass, mechanical stroke limits, stiffness/damping adjustability, mounting space, and compatibility with the force/torque sensor. These constraints must be considered so that the passive interface improves contact robustness without reducing positioning accuracy or exceeding the robot payload and workspace limits.

Based on these assumptions, the governing dynamic equation for each end-effector configuration may be written as follows [19].

- No Passive Element

$$M_O\ddot{\mathbf{p}} + C_O\dot{\mathbf{p}} + \mathbf{G}_O + \mathbf{F}_f + \mathbf{F}_d = \mathbf{F} - \mathbf{F}_{ext} \tag{5}$$

- Spring-Only

$$M_O\ddot{\mathbf{p}} + C_O\dot{\mathbf{p}} + \mathbf{G}_O + \mathbf{F}_f + \mathbf{F}_d + K\mathbf{e} = \mathbf{F} - \mathbf{F}_{ext} \tag{6}$$

- Spring-Damper

$$M_O\ddot{\mathbf{p}} + C_O\dot{\mathbf{p}} + \mathbf{G}_O + \mathbf{F}_f + \mathbf{F}_d + B\dot{\mathbf{e}} + K\mathbf{e} = \mathbf{F} - \mathbf{F}_{ext} \tag{7}$$

where, $\mathbf{e}$ and $\dot{\mathbf{e}}$ are defined as,

$$\begin{aligned} \mathbf{e} &= \mathbf{p}_{eff} - \mathbf{p}_{env}, \\ \dot{\mathbf{e}} &= \dot{\mathbf{p}}_{eff} - \dot{\mathbf{p}}_{env}. \end{aligned} \tag{8}$$

with , $\mathbf{e} = \dot{\mathbf{e}} = \mathbf{0}$ for the case of free movement. Here, $\mathbf{p}_{eff}$ is the robot's end-effector position, and $\mathbf{p}_{env}$ denotes the equilibrium position of the environment (e.g., the wall's surface)

### C. Environment Model

To evaluate the robustness of the proposed hybrid position-force control scheme with passive degrees of freedom (DOF), the environment model incorporates both static and dynamic mechanical properties. This allows for a systematic assessment of the passive elements' effectiveness in mitigating shocks and enhancing stability under varying conditions. The modeling begins with a simple rigid wall in the simulation, progressively introducing elastic and damping behaviors, and culminates in a fully variable mechanical profile. Analyses are conducted across the three end-effector configurations, no passive DOF (rigid), spring-only, and spring-damper, to compare the robot's and controller's behavior with and without compliant elements.

Initially, the environment is modeled as a rigid wall, representing a baseline scenario with infinite stiffness and no compliance. To introduce static mechanical properties, elastic behavior is added to this rigid wall, simulating a compliant surface where deformation occurs upon contact. This is achieved by incorporating an environmental stiffness coefficient $K_{env}$, which governs the linear force response proportional to penetration depth. The force at the interface is thus expressed as $\mathbf{F}_{env} = K_{env}(\mathbf{p}_{eff} - \mathbf{p}_{env})$.

Next, dynamic properties are integrated by adding a damping coefficient $D_{env}$ to the elastic model, creating a viscoelastic environment akin to the Kelvin-Voigt model (parallel spring-damper). Here, $D_{env}$ is selected to represent moderate energy dissipation, with the total environmental force given by

$$\mathbf{F}_{env} = K_{env}(\mathbf{p}_{eff} - \mathbf{p}_{env}) + D_{env}(\dot{\mathbf{p}}_{eff} - \dot{\mathbf{p}}_{env}), \tag{9}$$

where $\dot{\mathbf{p}}_{eff}$ and $\dot{\mathbf{p}}_{env}$ are the respective velocities (assuming $\dot{\mathbf{p}}_{env} = \mathbf{0}$ for a static wall). This formulation captures rate-dependent responses, such as those in vibration-prone or soft material interactions. Comparative simulations reveal that, in the absence of passive elements, the controller struggles with oscillatory responses and prolonged settling times during contact with this damped-elastic environment.

Finally, to represent repeatable time-varying interaction conditions in unstructured environments, a sinusoidal normal force disturbance is introduced at the contact interface. This disturbance is intended to approximate bounded periodic variations in contact loading that may arise from moving or deformable surfaces, such as slow human-body motion, breathing-like motion, compliant tissue deformation, or periodic vibration of the contacted environment. The sinusoidal model is not intended to reproduce a specific experimentally measured biological signal such as heartbeat or respiration. Instead, it is used as a representative bounded periodic excitation that allows the three end-effector configurations to be compared under identical and repeatable time-varying contact conditions. The time-varying interaction condition is implemented through the following normal force input:

$$\mathbf{F}_{env} = \left[ A\sin\left(\frac{2\pi}{T}t + \phi\right) + B \quad 0 \right]^T. \tag{10}$$

In this formulation, $A$ defines the disturbance amplitude, $B$ defines the mean contact-load offset, $T$ defines the disturbance period, and $\phi$ defines the initial phase. The parameter values were selected to produce a bounded and repeatable contact excitation over the simulation interval, rather than to match a specific experimentally measured biological or material response. The period $T$ was selected to generate observable transient and periodic effects during the simulation interval, allowing the controller and passive interface to be compared under repeatable time-varying contact excitation.

In this study, the constrained and unconstrained task-space directions are determined from the geometry of the contact surface. Since the environment is modeled as a flat wall, the surface normal defines the constrained direction, along which the contact force must be regulated. The tangential direction along the wall surface is treated as the unconstrained direction, along which the end-effector position trajectory is tracked. Therefore, the normal axis is assigned to force control, while the tangential axis is assigned to position control.

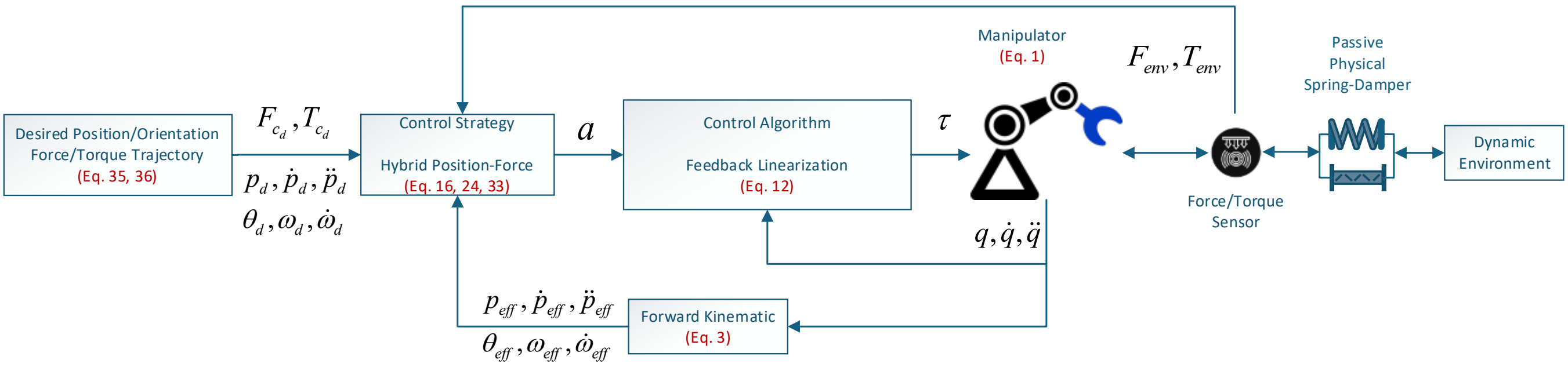

Fig.1: Block diagram of the proposed hybrid position-force control scheme with feedback linearization and integrated passive compliant degrees of freedom for robot-environment interaction.

## D. Overview of the Control Architecture

The overall architecture of the closed-loop control system is depicted in Fig. 1. It incorporates the specified motion and force trajectories, represented as $\mathbf{p}_D(t)$ and $\mathbf{F}_D(t)$. The hybrid force-position control scheme receives as inputs the desired references along with the actual end-effector position, which is computed via the forward dynamic module employing the Jacobian matrix to derive the robot's kinematic parameters in task space. Additionally, the reaction force arising from the robot's interaction with the environment serves as a key input to the feedback linearization controller, whose output consists of the actuator torques applied to the robot's joints.

position control scheme for a manipulator. The dynamic Eq. (1) can be reformulated using the workspace acceleration, disturbance, and friction force Eq. (3), yielding,

$$M(\mathbf{q})J^{-1}(\ddot{\mathbf{p}} - \dot{J}\dot{\mathbf{q}}) + C(\mathbf{q},\dot{\mathbf{q}})\dot{\mathbf{q}} + \mathbf{G}(\mathbf{q}) + \boldsymbol{\tau}_f(\dot{\mathbf{q}}) + \boldsymbol{\tau}_d = \boldsymbol{\tau} - J^T\mathbf{F}_{ext}. \quad (11)$$

The feedback-linearization design is developed under the standard assumptions of exact knowledge of the robot dynamic model, availability of joint position and velocity measurements, maintained contact during force regulation, and a nonsingular Jacobian along the considered trajectory. Under these nominal assumptions, the nonlinear robot dynamics can be transformed into a decoupled task-space acceleration model, which allows independent design of the position and force error dynamics. The stability result presented below therefore applies to the nominal closed-loop model; robustness to contact uncertainty is evaluated numerically through the passive compliant interface and the fixed/variable environment simulations.

The feedback-linearization controller assumes that the Jacobian remains nonsingular or sufficiently well-conditioned along the selected trajectory. In the simulations, the desired end-effector trajectory was selected within the reachable workspace of the 2-DOF manipulator and away from fully stretched or folded configurations, which are the main singular configurations of a planar two-link robot. The joint trajectories were monitored during simulation to ensure that the determinant of the Jacobian did not approach zero and that the inverse or pseudo-inverse mapping remained numerically well-conditioned.

Feedback linearization, based on the dynamic Eq. (3), can be expressed as,

$$M(\mathbf{q})J^{-1}(\mathbf{a} - \dot{J}\dot{\mathbf{q}}) + C(\mathbf{q},\dot{\mathbf{q}})\dot{\mathbf{q}} + \mathbf{G}(\mathbf{q}) + \boldsymbol{\tau}_f(\dot{\mathbf{q}}) + \boldsymbol{\tau}_d = \boldsymbol{\tau} - J^T\mathbf{F}_{ext}, \quad (12)$$

The system encompasses a Simscape-based robot model and an environmental object with fixed or time-varying interaction properties, represented by a flat contact surface approached by the robot during operation. For practical implementation, the required measurements include joint positions and velocities from standard robot sensors and interaction forces from an end-effector force/torque sensor, while the passive spring-damper element can be physically mounted between the robot wrist and the contact tool. Control Method

## E. Feedback Linearization for Motion Regulation

The block diagram of the proposed control architecture is illustrated in the Fig. 1, which depicts the hybrid force-where $\mathbf{a}$ represents a vector for the linear control strategy governing both force and position. Substituting Eq. (12) into Eq. (11) results in,

$$\ddot{\mathbf{p}} = \mathbf{a}. \quad (13)$$

Equation (13) shows that, under exact cancellation of the nonlinear dynamics, each controlled task-space direction behaves as a double-integrator system driven by the auxiliary input $\mathbf{a}$. The subsequent position and force controllers are therefore selected so that the corresponding tracking errors satisfy second-order homogeneous linear differential equations with positive feedback gains. This provides nominal asymptotic stability of the tracking errors.

The stability analysis presented in this section assumes ideal torque actuation, meaning that the commanded joint torque $\boldsymbol{\tau}$ can be delivered without saturation and with negligible delay. Under actuator saturation, exact feedback linearization may no longer be achieved because the applied torque differs from the commanded torque. In that case, the nominal error dynamics in Eqs. (17), (25), and (34) may be disturbed, and stability cannot be guaranteed solely by the nominal gain selection.

In practical implementations, actuator torque limits should therefore be included in the control design. Possible approaches include torque saturation compensation, anti-windup mechanisms, gain scheduling, constrained optimization, or model predictive control. In the present study, actuator saturation is not explicitly modeled; however, the joint-torque responses are reported to evaluate whether the passive compliant interface reduces torque fluctuations and peak control effort in the simulated scenarios.

Near singular configurations, the feedback-linearization law may generate large joint torques because small task-space motions can require large joint-space velocities or accelerations. Therefore, the nominal stability result is valid only when the Jacobian is nonsingular and the required control torques remain within actuator limits. If the robot operates close to singularities, additional measures such as damped least-squares inversion, singularity avoidance, trajectory replanning, or torque saturation handling are required.

It can be observed that the motion in the workspace is fully linearized and decoupled, allowing for the independent design of position and force controllers. In a specific scenario, position controllers can be tailored for workspace variables representing tangential motion, while force controllers can address those corresponding to normal forces.

*F. Hybrid Position-Force Control Framework*

The hybrid control decomposition is based on the known local contact frame. For the planar wall interaction considered in this work, the local contact frame is defined by the normal vector $\boldsymbol{n}$ to the wall and the tangential vector $\boldsymbol{t}$ along the wall surface. A selection matrix assigns the constrained normal direction to force control and the unconstrained tangential direction to position control. In the present 2-D case, this corresponds to regulating the normal contact force along the x-direction and tracking the tangential motion along the y-direction.

Given that the dynamics in the workspace are decoupled, the position, orientation, and force components are denoted accordingly. The position components in the workspace are given by

$$\ddot{\mathbf{p}}_T = \mathbf{a}_p, \qquad (14)$$

where $\mathbf{a}_p$ denotes the linear component of the position controller, and $\ddot{\mathbf{p}}_T$ represent the tangential accelerations of the end-effector on the surface. For feedback control purposes, the position tracking error is defined as,

$$\Delta\ddot{\mathbf{p}} = \ddot{\mathbf{p}}_{T_D} - \ddot{\mathbf{p}}_{T_{eff}}, \qquad (15)$$

Here $\ddot{\mathbf{p}}_{T_D}$ and $\ddot{\mathbf{p}}_{T_{eff}}$ denote the desired tangential acceleration trajectory relative to the environmental surface and the tangential acceleration components of the end-effector within the workspace, respectively. The associated linear controller is formulated as,

$$\mathbf{a}_p = \ddot{\mathbf{p}}_{T_D} + K_{T_V}\Delta\dot{\mathbf{p}} + K_{T_P}\Delta\mathbf{p}, \qquad (16)$$

with positive control gains $K_{T_V}$ and $K_{T_P}$. Substituting this into Eq. (12) produces the position tracking error dynamics as,

$$\Delta\ddot{\mathbf{p}} + K_{T_V}\Delta\dot{\mathbf{p}} + K_{T_P}\Delta\mathbf{p} = \mathbf{0}. \qquad (17)$$

Since $K_{T_V}$ and $K_{T_P}$ are selected as positive definite gain matrices, the characteristic equation of the position-error dynamics is Hurwitz; therefore, under the nominal feedback-linearized model, the tangential position-tracking error asymptotically converges to zero.

$$\lim_{t\to\infty}\Delta\mathbf{p} = \mathbf{0}. \qquad (18)$$

For force control in the direction normal to the environmental surface, the following relation is utilized as

$$\ddot{\mathbf{p}}_N = \mathbf{a}_N, \qquad (19)$$

where $\mathbf{a}_N$ is the linear component of the force controller normal to the surface, and $\ddot{\mathbf{p}}_N$ is the normal acceleration of the end-effector on the surface. Assuming the environment is modeled as a spring, the normal force applied to the environment is obtained from Eq. (20).

$$\mathbf{f}_{N_{eff}} = K_{env}(\mathbf{p}_{N_{eff}} - \mathbf{p}_{env}) \qquad (20)$$

where, $K_{env}$ is the environmental stiffness matrix, and $\mathbf{p}_{env}$ denotes the environmental position in the normal direction. Taking the second time derivative of Eq. (20) gives,

$$\ddot{\mathbf{p}}_{N_{eff}} = \frac{1}{K_{env}}\ddot{\mathbf{f}}_{N_{eff}}. \qquad (21)$$

This expresses the normal acceleration in the workspace in terms of the second derivative of the normal force. Substituting Eq. (21) into Eq. (19) yields the dynamic forces as,

$$\mathbf{a}_N = \frac{1}{K_{env}}\ddot{\mathbf{f}}_{N_{eff}}, \qquad (22)$$

For feedback control, the force tracking error is defined as,

$$\Delta\mathbf{f} = \mathbf{f}_{N_D} - \mathbf{f}_{N_{eff}}, \qquad (23)$$

where $\mathbf{f}_{N_D}$ is the desired normal force exerted on the environmental surface. The associated linear controller is,

$$\mathbf{a}_N = \frac{1}{K_{env}}(\ddot{\mathbf{f}}_{N_D} + K_{N_V}\Delta\dot{\mathbf{f}} + K_{N_P}\Delta\mathbf{f}), \qquad (24)$$

with a positive control gain matrix $K_{N_V}$ and $K_{N_P}$. Substituting this into Eq. (22) results in the following force tracking error dynamics.

$$\Delta\ddot{\mathbf{f}} + K_{N_V}\Delta\dot{\mathbf{f}} + K_{N_P}\Delta\mathbf{f} = \mathbf{0}, \qquad (25)$$

Since $K_{N_V}$ and $K_{N_P}$ are selected as positive definite gain matrices, the force-error dynamics in Eq. (25) are Hurwitz under the nominal contact model; therefore, the normal force-tracking error asymptotically converges to zero in the ideal feedback-linearized system.

$$\lim_{t\to\infty}\Delta\mathbf{f} = \mathbf{0}. \qquad (26)$$

This separation is valid as long as the contact geometry is known or can be estimated reliably and the surface normal does not change abruptly. Under severe uncertainty, such as unknown surface geometry, rapidly changing contact normals, contact loss, or multi-point contact, the fixed normal-tangential decomposition may no longer remain valid. In such cases, online estimation of the contact frame, adaptive selection matrices, or impedance/admittance-based transition strategies would be required. The present study focuses on a known flat-surface interaction so that the effect of the passive compliant DOF can be isolated and compared across the three end-effector configurations.

For orientation control in the workspace, the following equation is employed.

$$\dot{\boldsymbol{\omega}}_{eff} = \mathbf{a}_O. \qquad (27)$$

The most common approach to defining the orientation error involves a term similar to position control as,

$$\Delta\boldsymbol{\phi} = \boldsymbol{\phi}_D - \boldsymbol{\phi}_{eff}, \qquad (28)$$

where $\boldsymbol{\phi}_D$ and $\boldsymbol{\phi}_{eff}$ are the Euler angles corresponding to the end-effector frame orientation and the desired trajectory frame, respectively. These angles are extracted from the rotation matrices associated with the end-effector and the desired path.

Since Eq. (27) includes this term, the relationship between the time derivatives of the Euler angles and the end-effector angular velocity is crucial, as given by,

$$\boldsymbol{\omega}_{eff} = T(\boldsymbol{\phi}_{eff})\dot{\boldsymbol{\phi}}_{eff}, \qquad (29)$$

where, $T(\boldsymbol{\phi}_{eff})$ is the transformation matrix dependent on the specific set of Euler angles. The time derivative of Eq. (29) produces the acceleration equation as,

$$\dot{\boldsymbol{\omega}}_{eff} = T(\boldsymbol{\phi}_{eff})\ddot{\boldsymbol{\phi}}_{eff} + \dot{T}(\boldsymbol{\phi}_{eff}, \dot{\boldsymbol{\phi}}_{eff})\dot{\boldsymbol{\phi}}_{eff}. \qquad (30)$$

Given that the desired end-effector orientation trajectory is expressed in Euler angles, the time derivative of $\boldsymbol{\phi}_D$ is calculated as,

$$\begin{aligned} \boldsymbol{\omega}_D &= T(\boldsymbol{\phi}_D)\dot{\boldsymbol{\phi}}_D, \\ \dot{\boldsymbol{\phi}}_D &= T^{-1}(\boldsymbol{\phi}_D)\boldsymbol{\omega}_D. \end{aligned} \qquad (31)$$

Meanwhile, the time derivative of the angular rates are obtained as,

$$\begin{aligned} \dot{\boldsymbol{\omega}}_D &= T(\boldsymbol{\phi}_D)\ddot{\boldsymbol{\phi}}_D + \dot{T}(\boldsymbol{\phi}_D, \dot{\boldsymbol{\phi}}_D)\dot{\boldsymbol{\phi}}_D, \\ \ddot{\boldsymbol{\phi}}_D &= T^{-1}(\boldsymbol{\phi}_D)(\dot{\boldsymbol{\omega}}_D - \dot{T}(\boldsymbol{\phi}_D, \dot{\boldsymbol{\phi}}_D)\dot{\boldsymbol{\phi}}_D). \end{aligned} \qquad (32)$$

The linear controller for the Euler angle accelerations can be selected as

$$\begin{aligned} \mathbf{a}_O = {} & T(\boldsymbol{\phi}_{eff})(\ddot{\boldsymbol{\phi}}_D + K_{O_V}\Delta\dot{\boldsymbol{\phi}} + K_{O_P}\Delta\boldsymbol{\phi}) \\ & + \dot{T}(\boldsymbol{\phi}_{eff}, \dot{\boldsymbol{\phi}}_{eff})\dot{\boldsymbol{\phi}}_{eff}, \end{aligned} \qquad (33)$$

where $K_{O_V}$ and $K_{O_P}$ are the feedback gain matrices. Substituting this into Eq. (27) yields the orientation tracking error as,

$$\Delta\ddot{\boldsymbol{\phi}} + K_{O_V}\Delta\dot{\boldsymbol{\phi}} + K_{O_P}\Delta\boldsymbol{\phi} = \mathbf{0}. \qquad (34)$$

Provided that $T(\boldsymbol{\phi}_{eff})$ remains nonsingular and $K_{O_V}$ and $K_{O_P}$ are positive definite, the orientation-error dynamics in Eq. (34) are exponentially stable under the nominal model.

It should be emphasized that the above stability argument is a nominal result based on exact feedback linearization, nonsingular kinematics, and the assumed contact model. In practical robot-environment interaction, modeling errors, force-sensor noise, actuator limits, discontinuous contact, and uncertainty in environmental stiffness or damping may prevent exact cancellation of the nonlinear dynamics. For this reason, the present work evaluates the passive compliant DOF as an additional physical mechanism for improving robustness against contact transients and model mismatch.

Although the simulations are performed on a 2-DOF planar robot, the control architecture is not limited to this specific manipulator. For robots with more degrees of freedom, the same principle can be extended by defining the task-space contact frame, assigning force control to constrained directions and position control to unconstrained directions, and mapping the resulting task-space control action to joint torques through the robot Jacobian. However, practical implementation on other robots requires recalculation of the dynamic model, Jacobian, actuator limits, and passive interface parameters for the specific robot structure.

## III. Numerical Simulation

This section outlines simulations and comparative analyses used to evaluate the feasibility and performance of the proposed approach.

### *A. Software and Tools*

All simulations are performed using MATLAB and the Simulink platform, where the environment interacts with the robot via external forces at the contact interface.

Although the present study is simulation-based, the proposed control structure is practically implementable in principle. The feedback-linearized hybrid controller requires the robot dynamic model, joint position and velocity measurements, and force feedback from the end-effector. These requirements are compatible with many industrial and collaborative robot platforms equipped with joint encoders and wrist force/torque sensors. The passive compliant DOF can be implemented as a mechanical spring-damper module placed between the wrist and the end-effector tool. However, practical implementation would require consideration of actuator torque limits, sensor noise, sampling rate, unmodeled friction, mechanical backlash, and possible saturation of the passive element.

Within the MATLAB/Simulink framework, this fluctuating environment is realized as a signal feeding into the Spatial contact Force block. The manipulator is assigned a path-following task while maintaining a prescribed end-effector contact force.

### *B. Scenarios*

Simulations were conducted across each end-effector configuration under fixed and time-varying interaction conditions in order to evaluate how changes in environmental stiffness, damping, and contact excitation affect the relative performance of the rigid, spring-only, and spring-damper interfaces. This methodology promotes enhanced safety and adaptability in robotic interactions, thereby facilitating broader implementation in unpredictable or unstructured environments.

The evaluated scenarios encompassed surface-approaching maneuvers under predefined force conditions followed by trajectory tracking tasks, for the selected end-effector and environment configurations.

### *C. Parameters*

The geometric and inertial parameters of the 2-DOF manipulator were chosen to represent a small-scale planar robotic arm suitable for contact-control simulation, with link lengths and masses listed in Appendix Table I. The control gain matrices were selected as diagonal positive definite matrices to obtain decoupled second-order error dynamics in the feedback-linearized task space. For the tangential position controller, $K_{T_P} = k_{T_P}\text{diag}([1, \dots, 1])$ and $K_{T_V} = k_{T_V}\text{diag}([1, \dots, 1])$ determine the nominal stiffness and damping of the position-error dynamics. For the normal force controller, $K_{N_P} = k_{N_P}\text{diag}([1, \dots, 1])$ and $K_{N_V} = k_{N_V}\text{diag}([1, \dots, 1])$ determine the convergence rate and damping of the force-error dynamics. For the orientation controller, $K_{O_P} = k_{O_P}\text{diag}([1, \dots, 1])$ and $K_{O_V} = k_{O_V}\text{diag}([1, \dots, 1])$ define the nominal orientation-error response. The scalar coefficients were initially chosen to ensure positive definite error dynamics and then adjusted empirically in simulation to obtain stable tracking without excessive oscillation or unrealistically large actuator torques. The final values are reported in Appendix Table II.

The tuning procedure followed a two-step approach. First, proportional gains were selected to obtain a sufficiently fast nominal response in the corresponding position, force, or orientation channel. Second, velocity/derivative gains were increased to provide damping and reduce overshoot. The gains were then refined through simulation to maintain stable contact behavior in both fixed and time-varying environment cases. Different normal-force gains were used in the fixed and variable cases because the variable environment includes a larger time-varying contact excitation, and lower force-loop gains helped avoid excessive force oscillations and torque demand.

Regarding the static mechanical attributes of the environment, the stiffness and damping coefficients are denoted as $K_{env}$ and $D_{env}$ , respectively. Under time-varying interaction conditions, as given in Eq. (10), a sinusoidal normal force disturbance with amplitude $A$, mean offset $B$, phase $\phi$, and period $T$ is applied at the contact interface. The amplitude $A$ determines the magnitude of the periodic load variation, $B$ represents the mean contact-load level, $\phi$ sets the initial phase, and $T$ determines the rate of variation over the simulation interval. These values were selected to create a bounded repeatable disturbance for comparative evaluation, not to reproduce a specific experimentally measured biological process. The selected values are reported in Appendix Table III.

Therefore, the time-varying environment case should be interpreted as a controlled robustness test against periodic contact-load variation rather than as a validated model of breathing, heartbeat, or a particular soft-tissue response.

The environmental stiffness, damping, and sinusoidal force parameters were selected as representative simulation values rather than experimentally identified parameters for a specific material. The fixed environment case uses a Kelvin-Voigt-type linear contact model to represent an elastic-damped surface. The variable environment case introduces a bounded sinusoidal normal force disturbance to emulate repeatable variations in contact loading, such as those that may occur in interaction with deformable or moving surfaces. Therefore, the model should be interpreted as a controlled simulation scenario for comparative evaluation, not as a validated biological or material-specific contact model.

The selected environmental stiffness and damping values were chosen to represent a moderate contact condition suitable for comparing the three end-effector configurations under identical simulation settings. Extremely high stiffness values were not used in the main simulations because they can introduce numerical stiffness in the contact solver and may require smaller integration time steps, higher actuator torques, or dedicated impact models. Extremely low stiffness values were also not emphasized because they reduce the severity of contact transients and may obscure the shock-mitigation role of the passive interface. Therefore, the present study focuses on a representative intermediate range to isolate the comparative effect of the passive DOF configurations.

The sensitivity of the system response to environmental stiffness and damping is important because these parameters are rarely known exactly in practical robot-environment interaction. A controller tuned for one contact condition may exhibit excessive overshoot, oscillation, or force-tracking error when the environment becomes softer, stiffer, or more dissipative. Therefore, the variation of environmental parameters was considered to assess whether the passive compliant DOF improves robustness under contact uncertainty rather than only under a single nominal case.

In the robotic setup, these environmental traits were complemented by end-effector mechanical features, including a configuration with only a spring and another incorporating both a spring and damper, with their respective parameters set as $k_{spring}$ and $D_{damping}$ , as outlined in Table IV.

The spring and damper coefficients were selected to represent physically plausible passive compliance levels for a small 2-DOF planar manipulator while preserving stable contact behavior in the simulated interaction tasks. The spring stiffness was chosen to provide sufficient compliance for impact attenuation without producing excessive end-effector deflection, whereas the damping coefficient was selected to dissipate the stored elastic energy and suppress rebound oscillations. Different values were used in the fixed and variable environment cases to account for the different contact excitation levels and simulation durations. These values are not claimed to be globally optimal; rather, they provide a representative parameter set for evaluating the relative influence of rigid, spring-only, and spring-damper end-effector configurations.

In a real industrial implementation, these passive parameters would need to be selected according to the robot payload, tool mass, allowable end-effector deflection, expected contact force range, and available mechanical stroke of the compliant module.

*D. Trajectories and Inputs*

A jerk-limited seven-segment trajectory, as referenced in [20], is employed for the two-degree-of-freedom (2-DOF) robot to ensure smooth motion planning with bounded jerk, acceleration, and velocity, thereby minimizing mechanical stress and vibrations during operation. The force input is defined as Eq. (35), applying a constant force of 10 N in the normal direction (along the x-axis) while maintaining zero force tangentially. The position trajectory is specified as Eq. (35), where the x-component remains fixed at 0.5 m, and the y-component follows the seven-segment profile in the tangential direction.

$$\begin{aligned} F &= [f_x, f_y]^T = [10,0]^T, \\ p &= [p_x, p_y]^T = [0.5,\text{7-Segment}]^T \end{aligned} \tag{35}$$

This seven-segment trajectory along the tangential direction ( $p_y$ ) is designed to produce controlled, trapezoidal velocity profiles with S-curve acceleration transitions, which incorporate phases of constant jerk to ramp up and down acceleration smoothly. The parameters governing this profile are as follows

$$\begin{aligned} &\bullet q_0 = 0, q_1 = 0.1, v_0 = 0, v_1 = 0, a_0 = 0, a_1 = 0 \\ &\bullet v_{min} = -0.2, v_{max} = 0.2 \\ &\bullet a_{min} = -0.01, a_{max} = +0.01 \\ &\bullet j_{min} = -0.005, j_{max} = +0.005 \\ &\bullet T_a = 22, T_v = -21.5, T_s = 0.001, T = 22.5 \end{aligned} \tag{36}$$

By applying a constant normal force ($f_x$) alongside the tangential position trajectory ($p_y$), the system achieves hybrid force-position control, where force regulates penetration or contact in one axis while position dictates sliding or following in the orthogonal axis. This configuration is scientifically sound for simulating realistic interactions, such as polishing or assembly on surfaces, as it prevents abrupt changes that could lead to instability or damage, with all bounds ensuring feasibility within typical actuator limits.

### *E. Simulation-Based Evaluation*

The simulation-based evaluation compares desired and actual end-effector trajectories, positions, velocities, forces, and joint torques across three 2-DOF robot configurations interacting with fixed and variable environment conditions, as illustrated in (Fig. 3 – Fig. 18) and summarized in (Table V – Table XIII). These configurations include a baseline robot without passive elements, one equipped with a spring at the end-effector, and another with both spring and damper. The first arm operates without passive elements, relying solely on active control to counteract environmental forces, which often leads to instability due to unmitigated momentum and energy transfer. The second arm incorporates a spring, enabling elastic energy storage according to Hooke's law ($F = -kx$, where $k$ is stiffness and $x$ is displacement), which helps absorb impacts but can induce rebounds if undamped. The third arm adds a damper, introducing viscous friction ($F = -b\dot{x}$, with $b$ as damping coefficient and $\dot{x}$ as velocity) to dissipate kinetic energy, promoting asymptotic stability. These scenarios were replicated in both static and varying environmental conditions as illustrated in Fig. 2:

- $S_1$: involved no passive elements.
- $S_2$: incorporated a spring.
- $S_3$: Equipped with the combination of a spring and a damper.

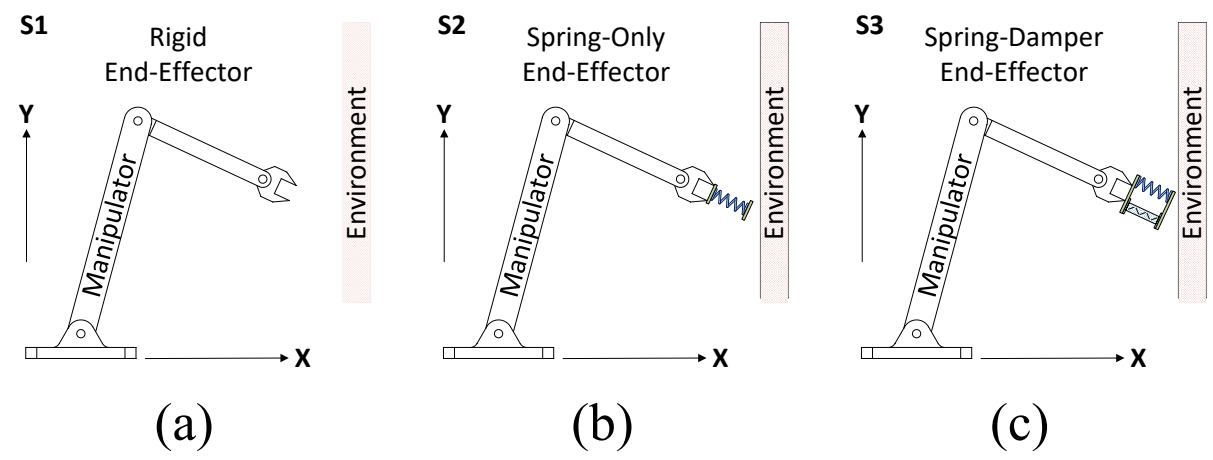

Fig.2: Two-link planar arm scenarios; (a): S1 (Rigid End-Effector); (b): S2 (End-Effector with spring only); (c): S3 (End-Effector with spring-damper).

In the fixed environment, where mechanical properties remain constant with stiffness and damping values unchanged, the model shows actual trajectories aligning closely with desired paths, particularly for the spring-damper equipped arm, as minimal deviations appear in the $X - Y$ plots over 0.35 to 0.5 meters in $X$ and 0 to 0.3 meters in $Y$, evident in the end-effector trajectory plot (Fig. 3). Error metrics for position in the position error plot (Fig. 5), velocity in the velocity error plot (Fig. 7), and force in the force error plot (Fig. 9) converge to near-zero values within two seconds for the spring-damper configuration, while the baseline and spring-only arms display persistent errors, reflecting the model's accuracy in simulating passive element effects on stability. This short settling time in fixed conditions indicates rapid achievement of steady state, with no further changes beyond two seconds, validating the simulation's efficiency in capturing equilibrium dynamics.

For the variable environment, where forces vary sinusoidally as $A\sin(\omega t) + B$, the model reliably reproduces oscillations over 22 seconds, with reduced error amplitudes in the spring-damper arm, as seen in the dynamic end-effector trajectory plot (Fig. 11), position error plot (Fig. 13), velocity error plot (Fig. 15), and force error plot (Fig. 17), showing trends that are consistent with established compliant-interaction behavior in robotic impedance control. Summary statistics from the static Table V- VIII further support this, showing progressive reductions in position error standard deviations from 0.0301 meters in the baseline arm's normal direction to 0.0277 meters in the spring-damper arm, and similar patterns in the dynamic summary Table IX- XII where velocity error standard deviations decrease from 0.186 meters per second in the baseline to 0.109 meters per second in the spring-damper, underscoring the model's reliability in both scenarios.

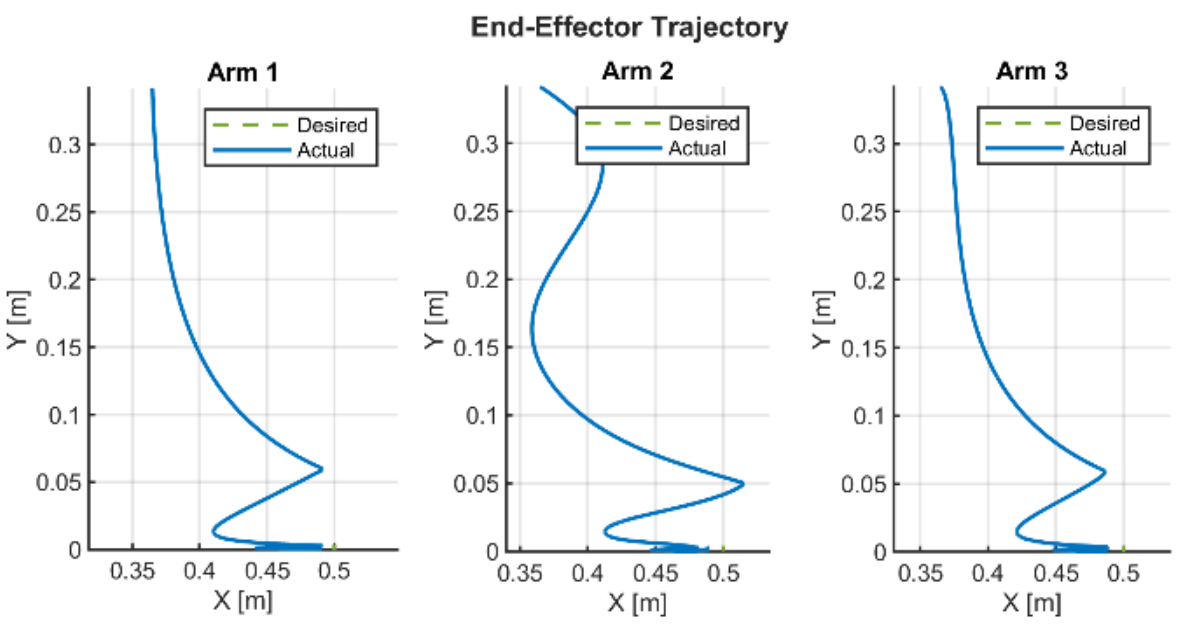

Figure 3: End-effector trajectories in the X-Y plane for the fixed environment interaction, showing desired (dashed) and actual (solid) paths for (a) baseline arm without passive elements (notable deviations due to oscillations), (b) arm with spring (reduced path wander via elastic compliance), and (c) arm with spring and damper (close agreement with the desired trajectory with minimal bounces, highlighting damping's stabilization effect).

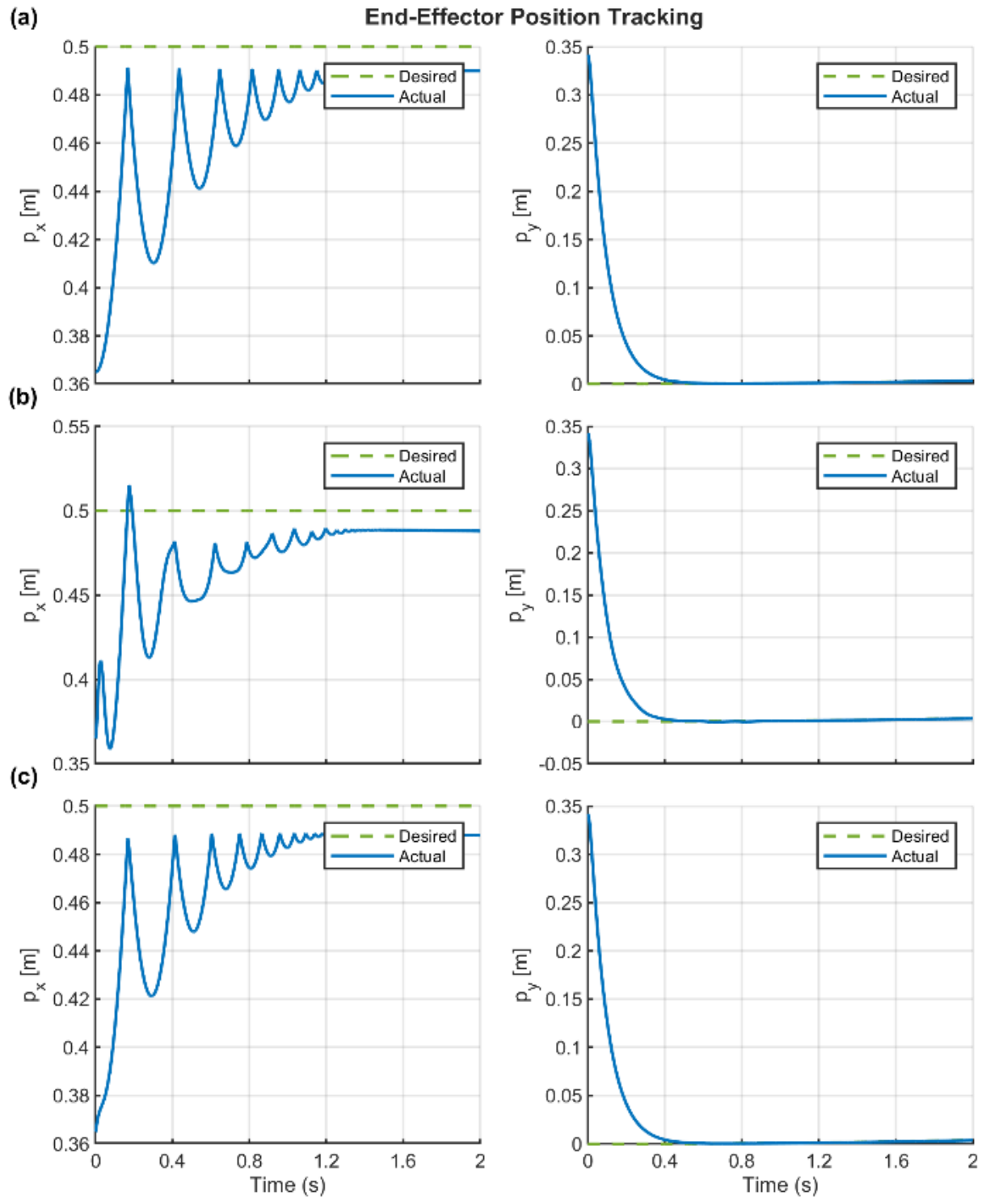

Figure 4: End-effector position tracking over time (0-2 s) in the fixed environment, depicting desired and actual positions in x (0.36-0.55 m) and y (0-0.4 m) directions for (a) baseline arm without passive elements (large overshoots up to 0.491 m normal), (b) arm with spring (moderated peaks but prolonged ringing), and (c) arm with spring and damper (rapid convergence to steady state ~0.5 m normal and 0.3 m tangential within 1 s).

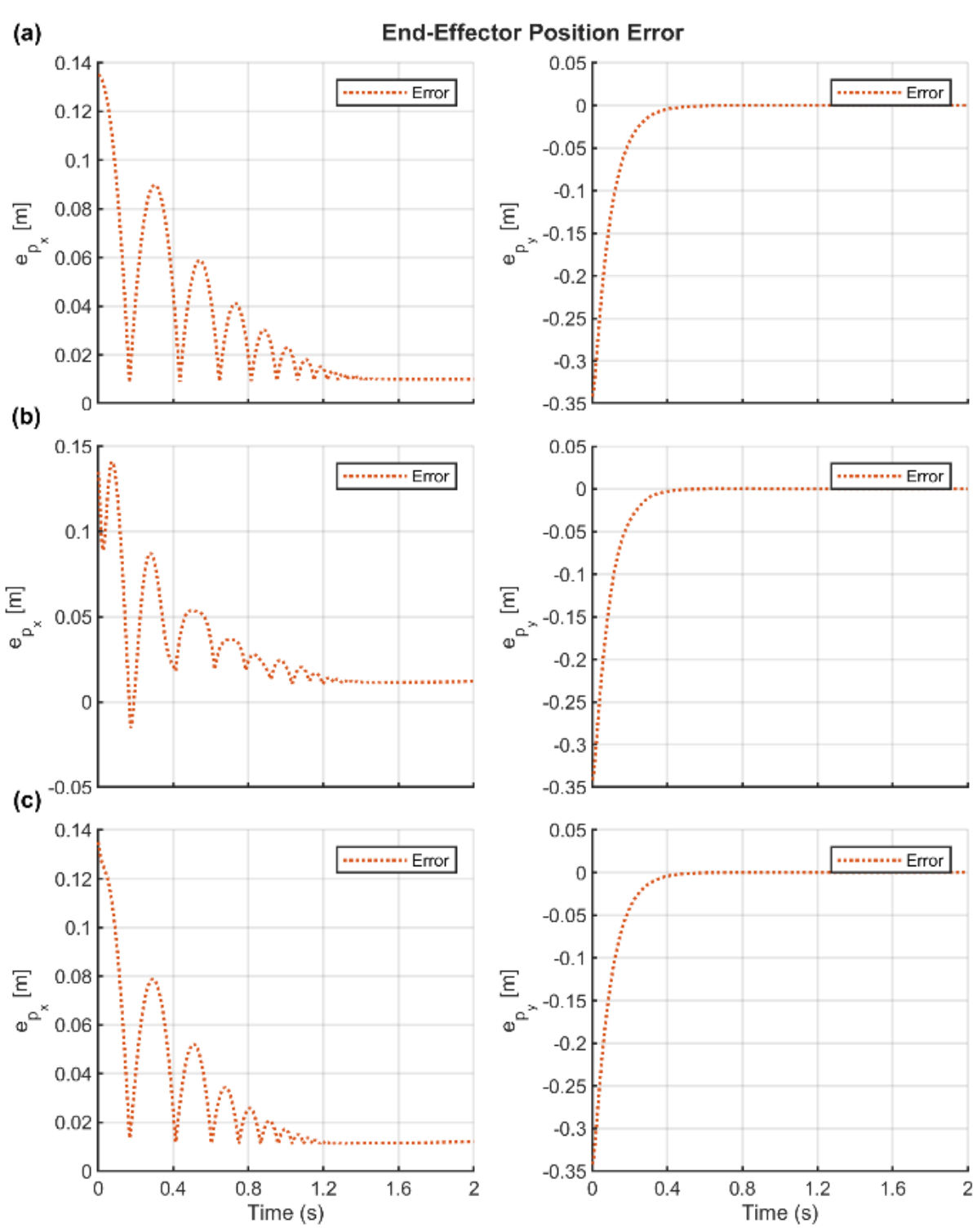

Figure 5: End-effector position errors in the fixed environment, illustrating error distributions in x (up to 0.15 m) and y (up to -0.4 m) for (a) baseline arm without passive elements (broad spreads from undamped resonances) and (b) arm with spring (narrowed but persistent), and time-series errors for (c) arm with spring and damper (convergence to zero, with std 0.0277 m normal highlighting vibration suppression).

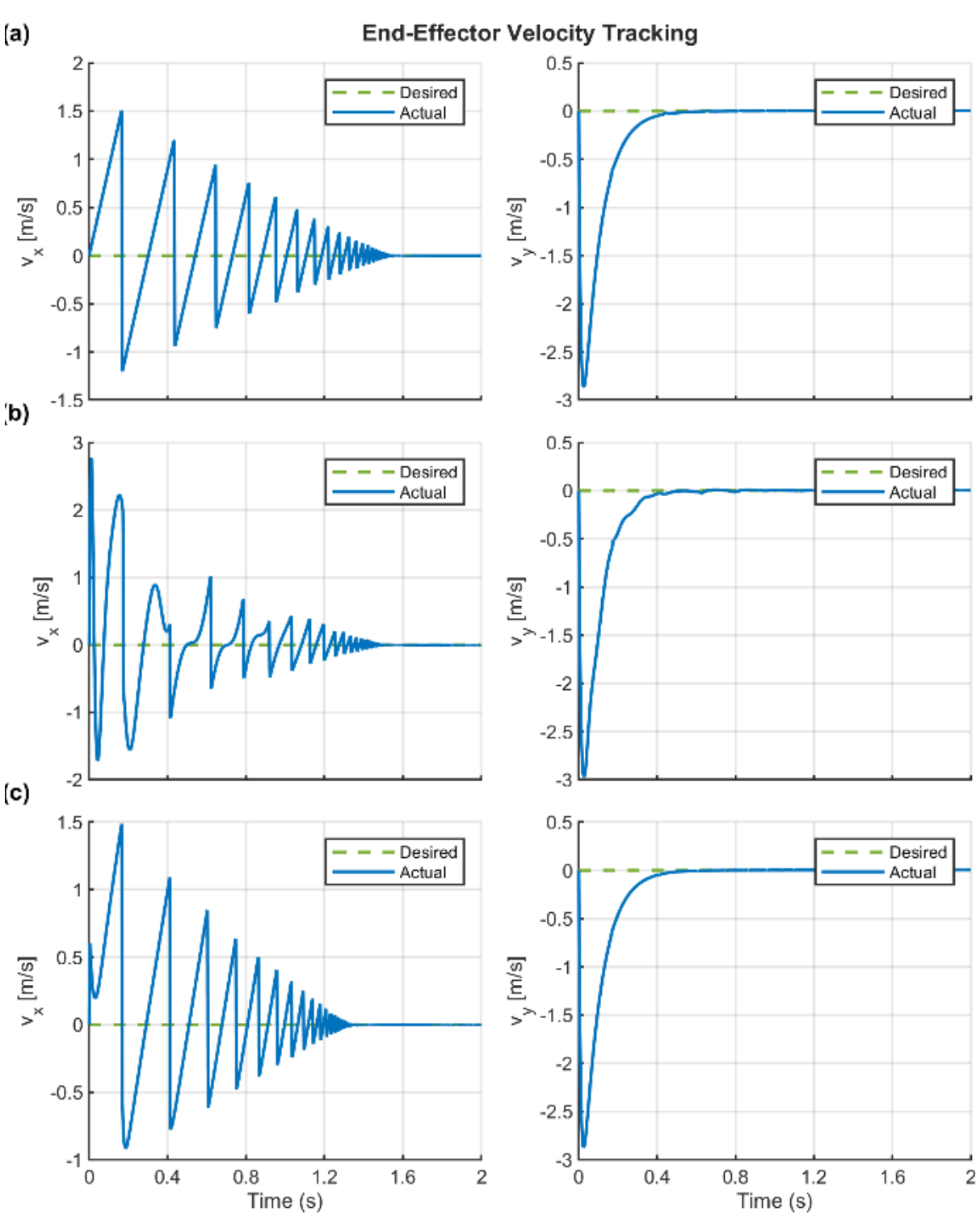

Figure 6: End-effector velocity tracking over time (0-2 s) in the fixed environment, showing desired and actual velocities in x (±2 m/s) and y (±3 m/s) directions for (a) baseline arm without passive elements (large velocity fluctuations from abrupt forces), (b) arm with spring (marginal improvement via energy storage), and (c) arm with spring and damper (stabilized within ±1 m/s, reducing jerk through viscous damping).

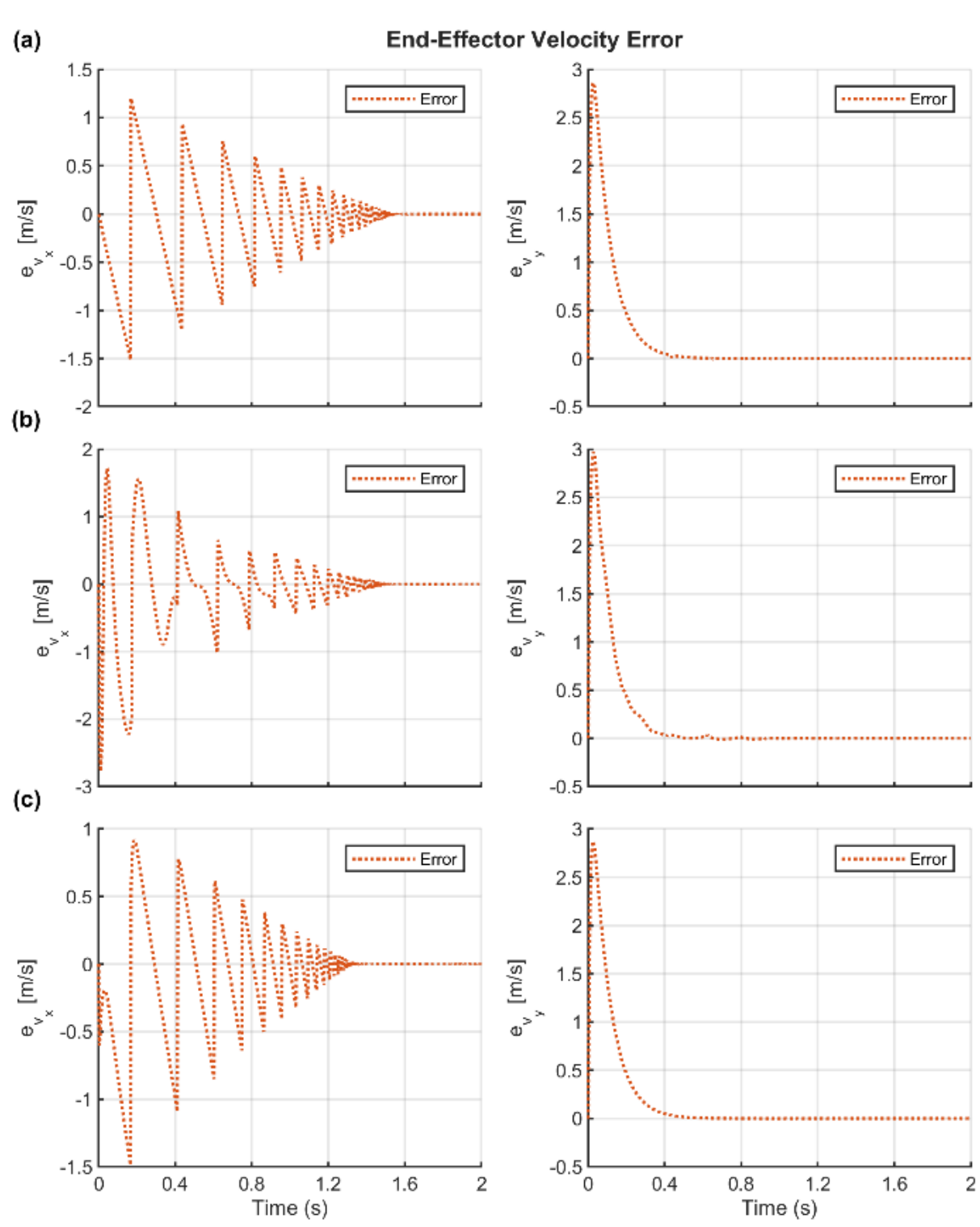

Figure 7: End-effector velocity errors in the fixed environment, displaying error distributions in x (±3 m/s) and y (±3 m/s) for (a) baseline arm without passive elements (peaks from phase lags) and (b) arm with spring (lessened but ongoing), and time-series errors for (c) arm with spring and damper (minimized to ±0.9 m/s, with std reductions ~13% normal emphasizing bandwidth extension).

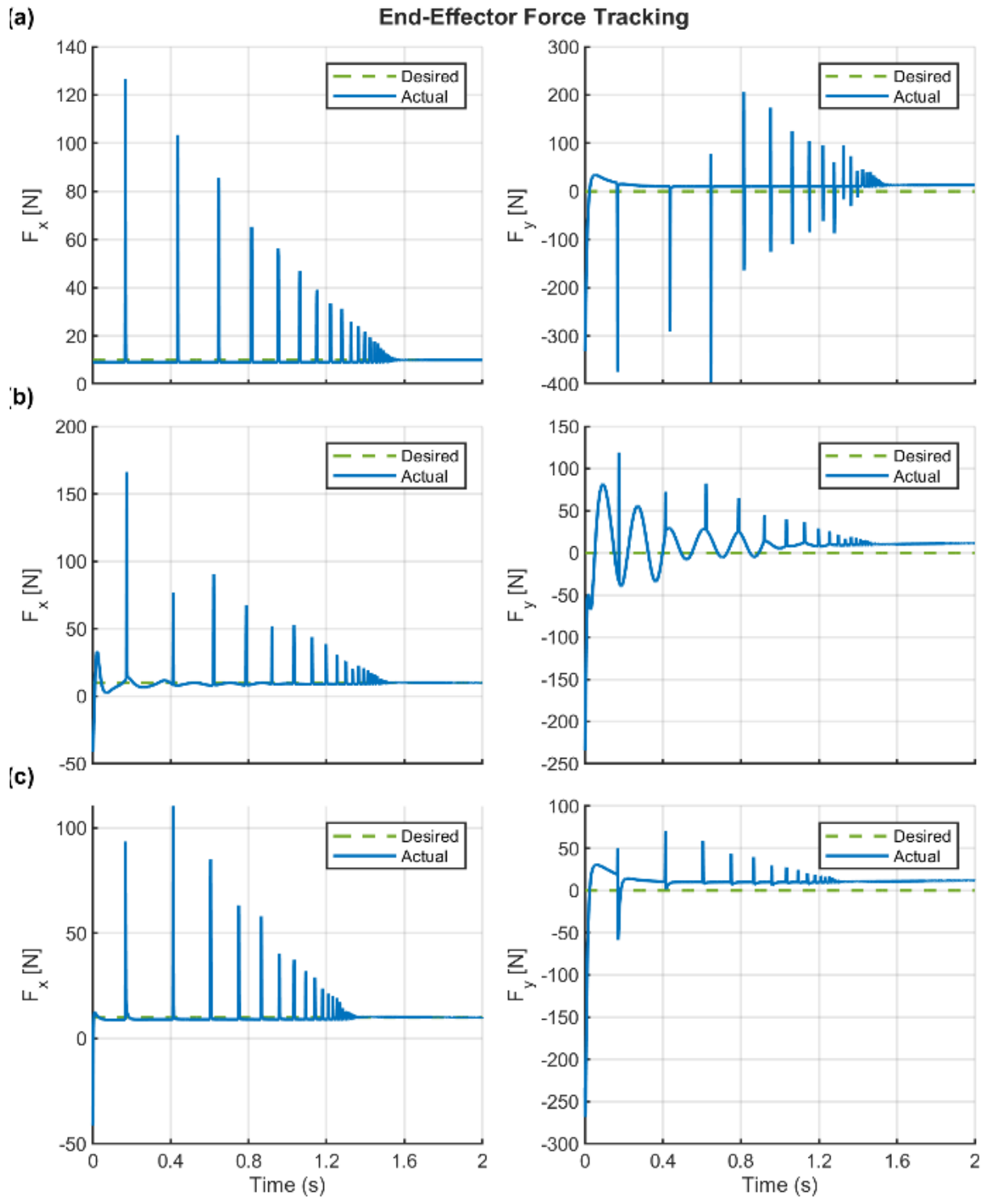

Figure 8: End-effector force tracking over time (0-2 s) in the fixed environment, presenting desired and actual forces in x (0-150 N) and y (-400-300 N) directions for (a) baseline arm without passive elements (overshoots to 127 N normal from stiff impacts), (b) arm with spring (smoothed transients), and (c) arm with spring and damper (tracked within 110 N, balancing via hybrid impedance).

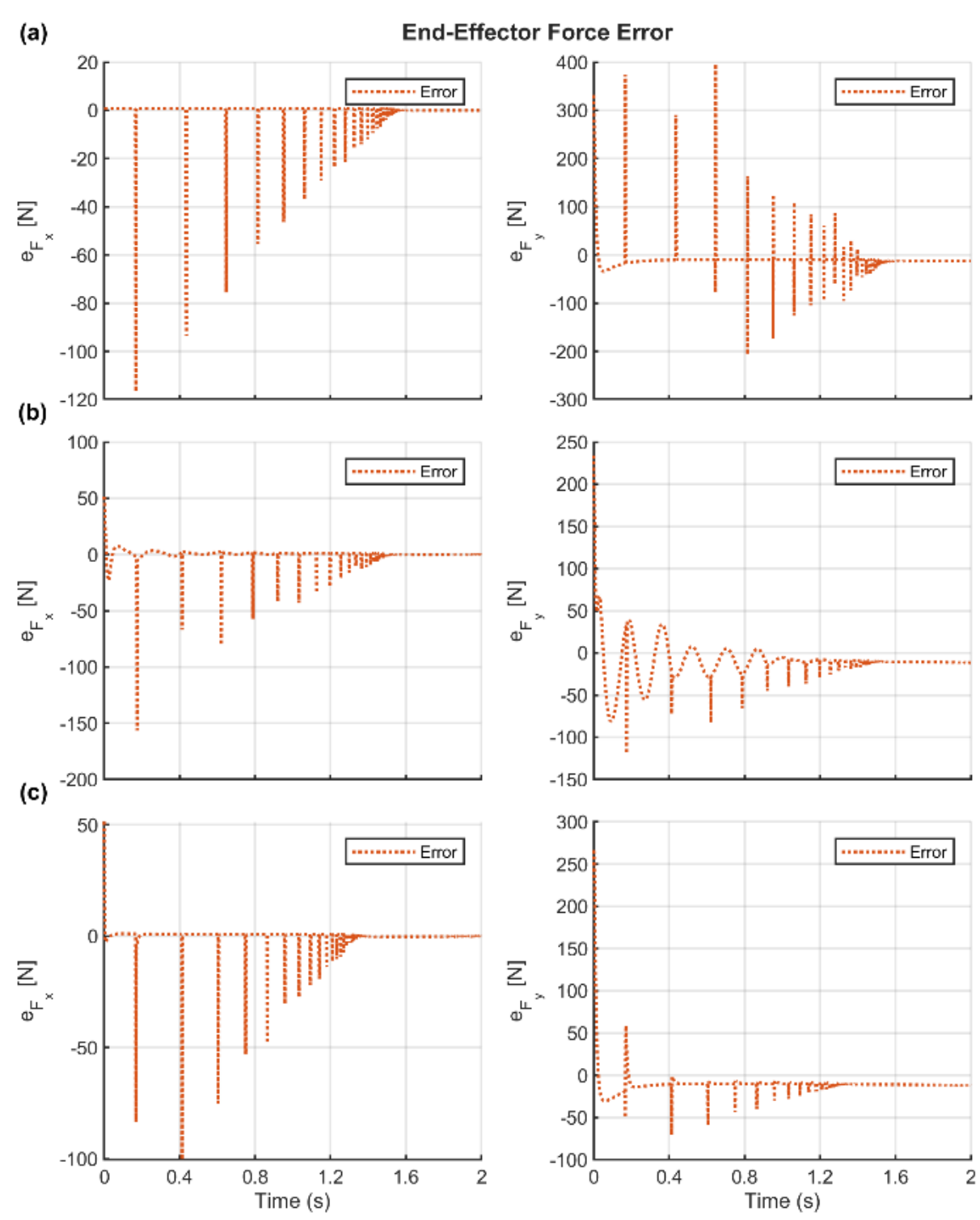

Figure 9: End-effector force errors in the fixed environment, showing error distributions in x (±100 N) and y (±400 N) for (a) baseline arm without passive elements (spreads up to 397 N tangential from unmitigated reactions) and (b) arm with spring (buffered), and time-series errors for (c) arm with spring and damper (damped to std 17.6 N tangential, attenuating ripples through energy dissipation).

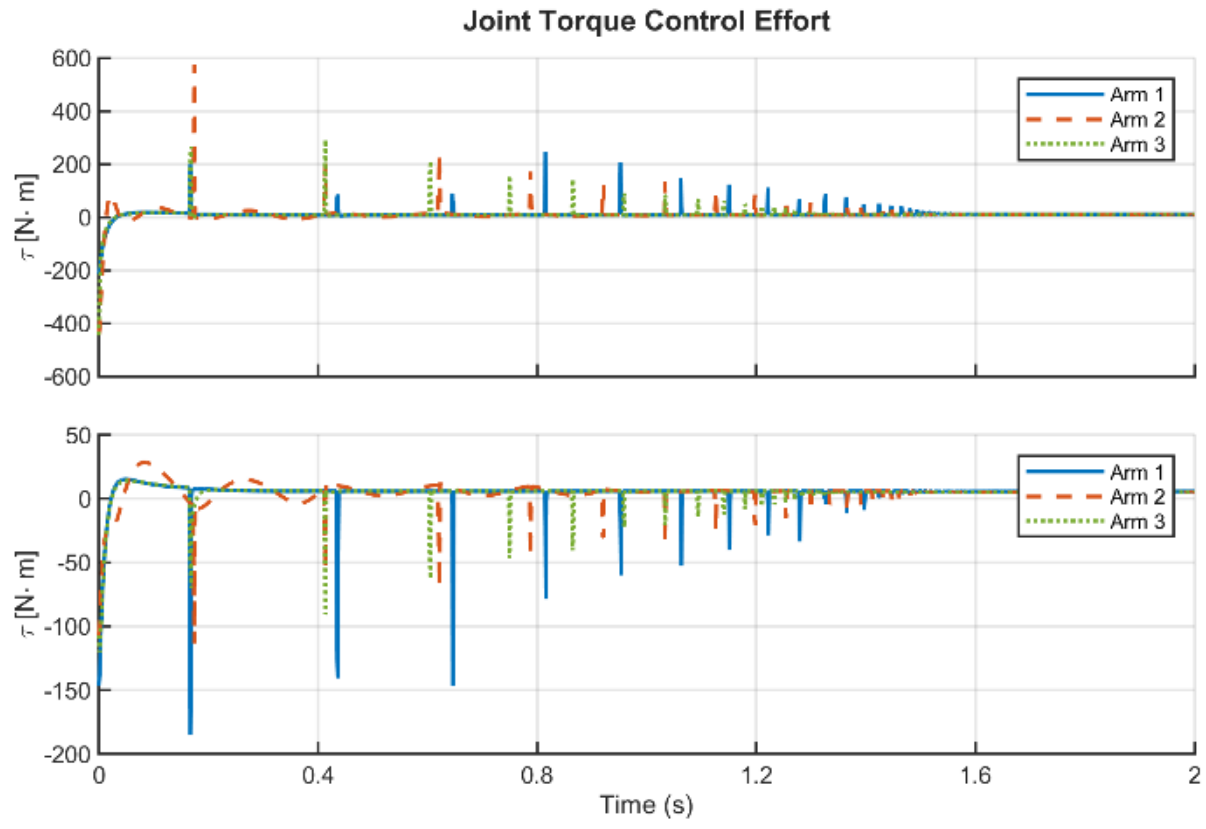

Figure 10: Joint torque control efforts over time (0-2 s) in the fixed environment, depicting torque time-series (±500 N·m) for Arm 1 (baseline without passive elements, peaks at 574 N·m from high gains), Arm 2 (with spring, moderated), and Arm 3 (with spring and damper, smoothed to std 21.7 N·m, indicating efficient actuator use).

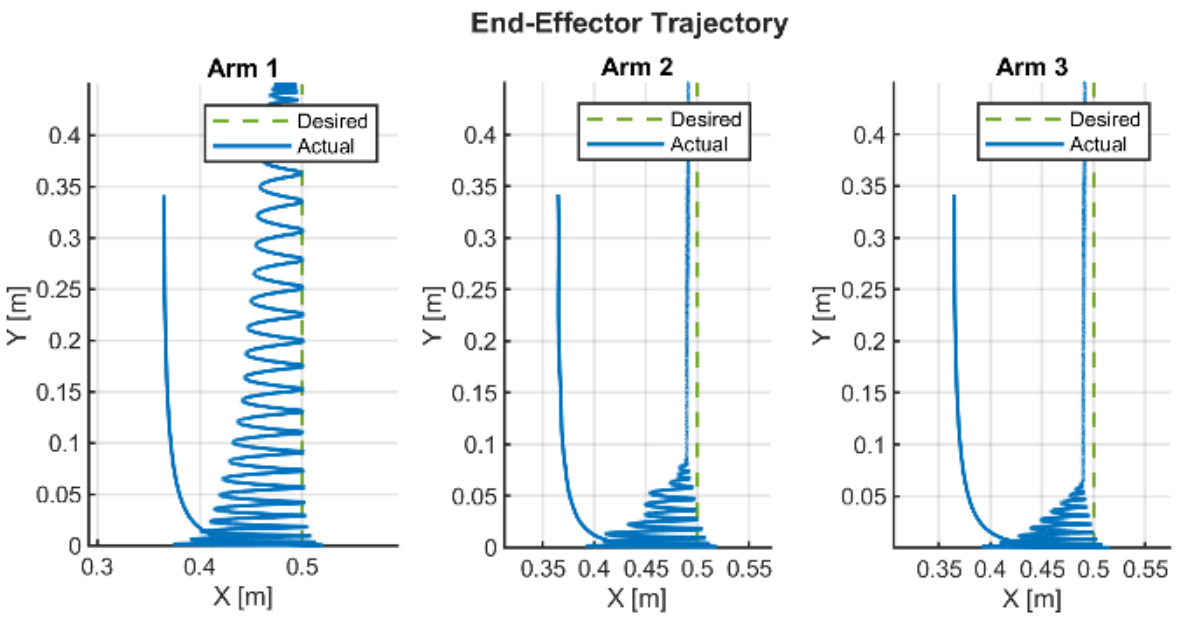

Figure 11: End-effector trajectories in the X-Y plane for the variable environment interaction over 22 s, showing desired (dashed) and actual (solid) paths for (a) baseline arm without passive elements (persistent oscillations diverging periodically), (b) arm with spring (amplitude mitigation), and (c) arm with spring and damper (suppressed resonances, with std 0.025 m normal highlighting adaptive compliance).

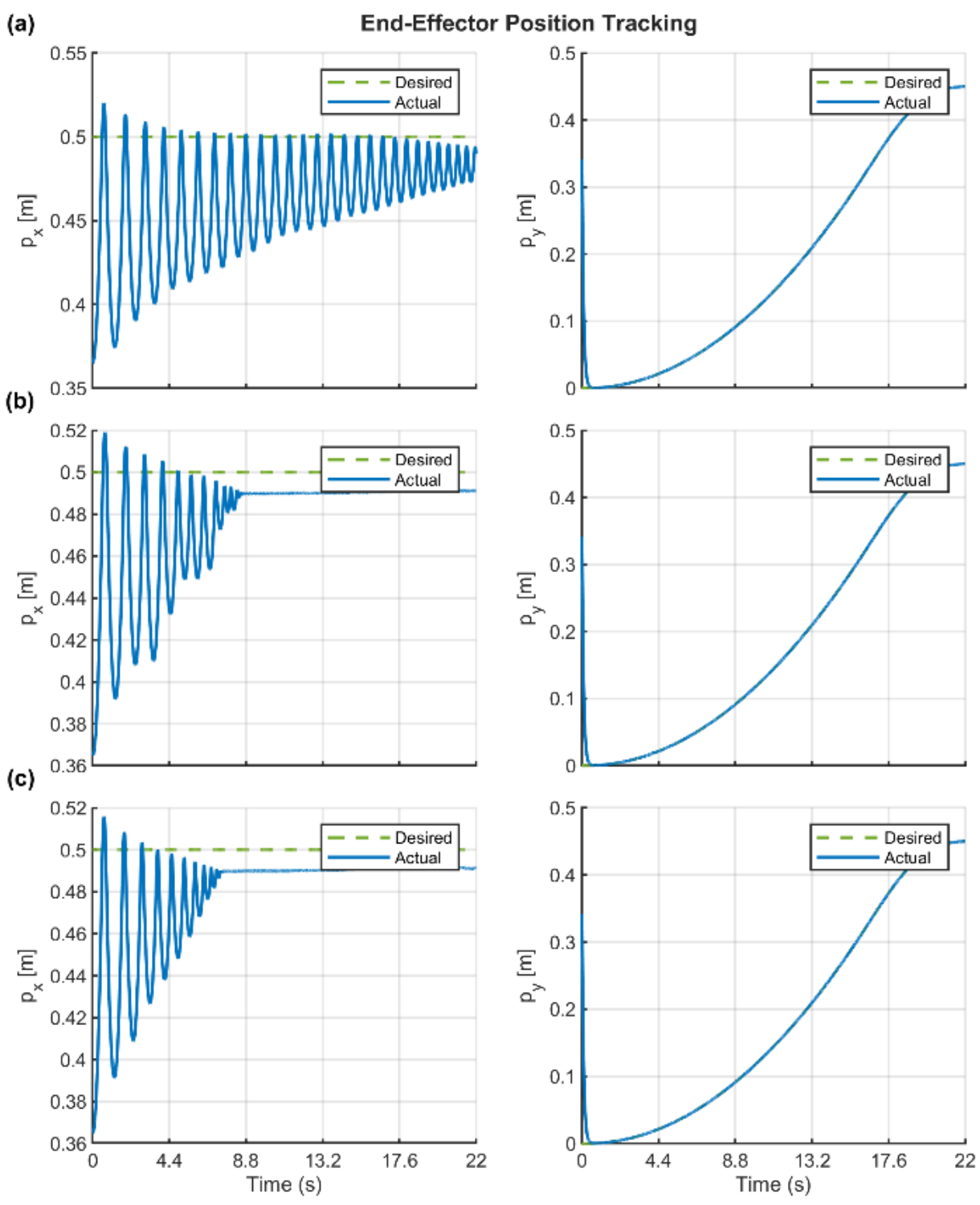

Figure 12: End-effector position tracking over time (0-22 s) in the variable environment, depicting desired and actual positions in x (0.36-0.55 m) and y (0-0.45 m) directions for (a) baseline arm without passive elements (overshoots to 0.520 m from excitations), (b) arm with spring (lessened

peaks), and (c) arm with spring and damper (minimal per-cycle deviations via critical damping).

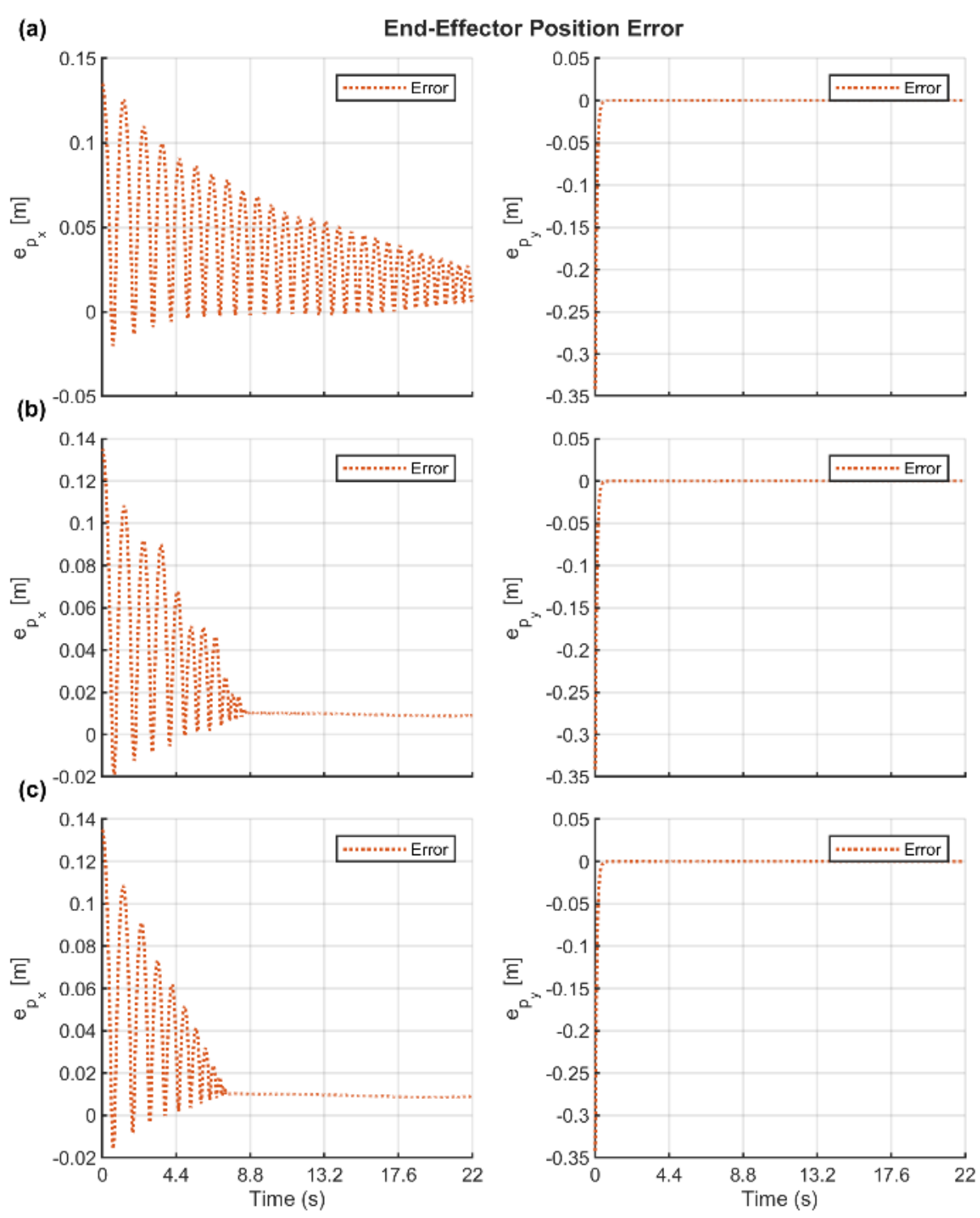

Figure 13: End-effector position errors in the variable environment, illustrating error distributions in x (up to 0.135 m) and y (up to -0.342 m) for (a) baseline arm without passive elements (wide oscillations, std 0.031 m normal) and (b) arm with spring (lowered peaks), and time-series errors for (c) arm with spring and damper (halved variances, embodying robustness against uncertainties).

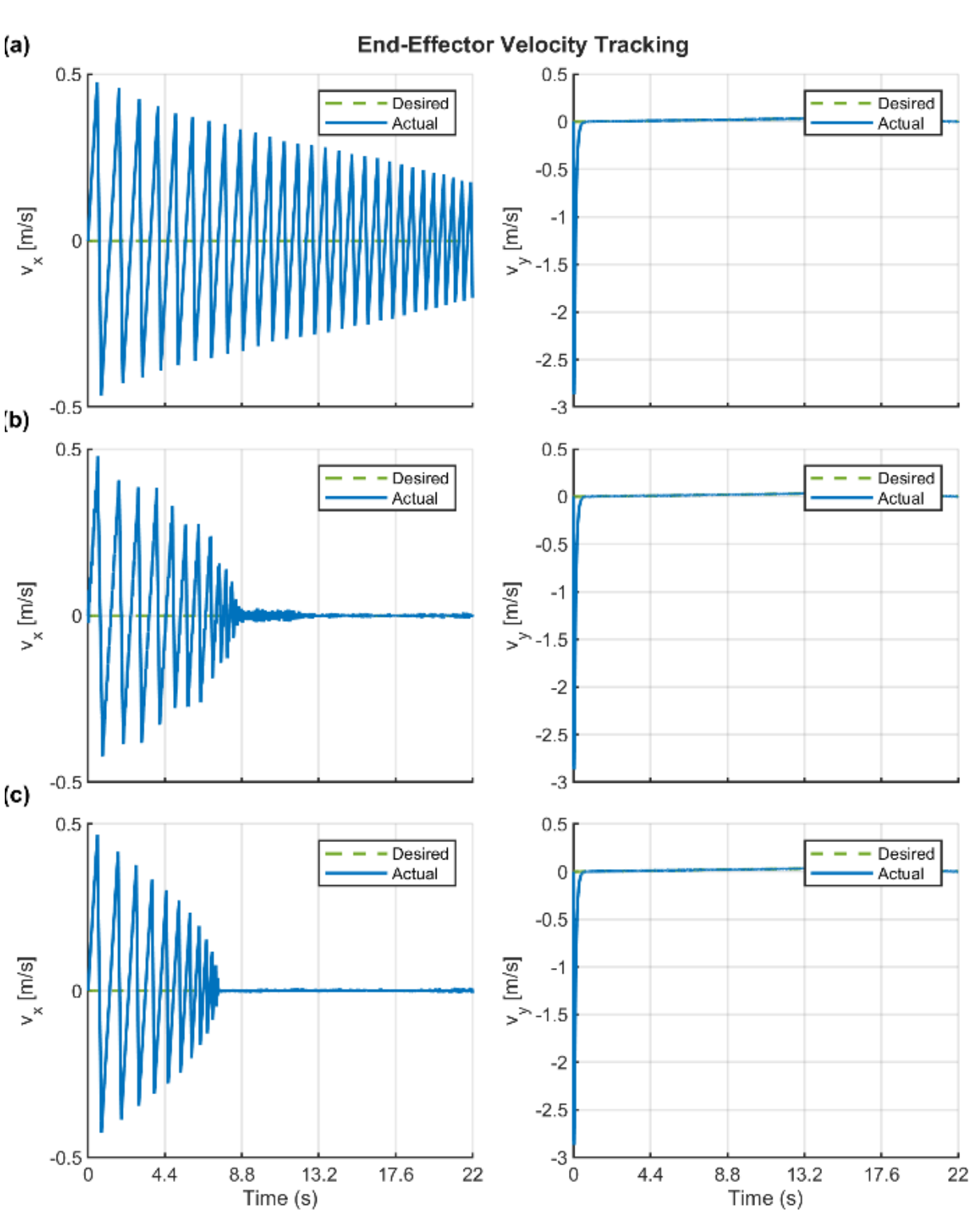

Figure 14: End-effector velocity tracking over time (0-22 s) in the variable environment, showing desired and actual velocities in x (±0.5 m/s) and y (±3 m/s) directions for (a) baseline arm without passive elements (swings to ±0.476 m/s from disruptions), (b) arm with spring (energy storage), and (c) arm with spring and damper (damped to ±0.426 m/s, with std reductions 41% normal).

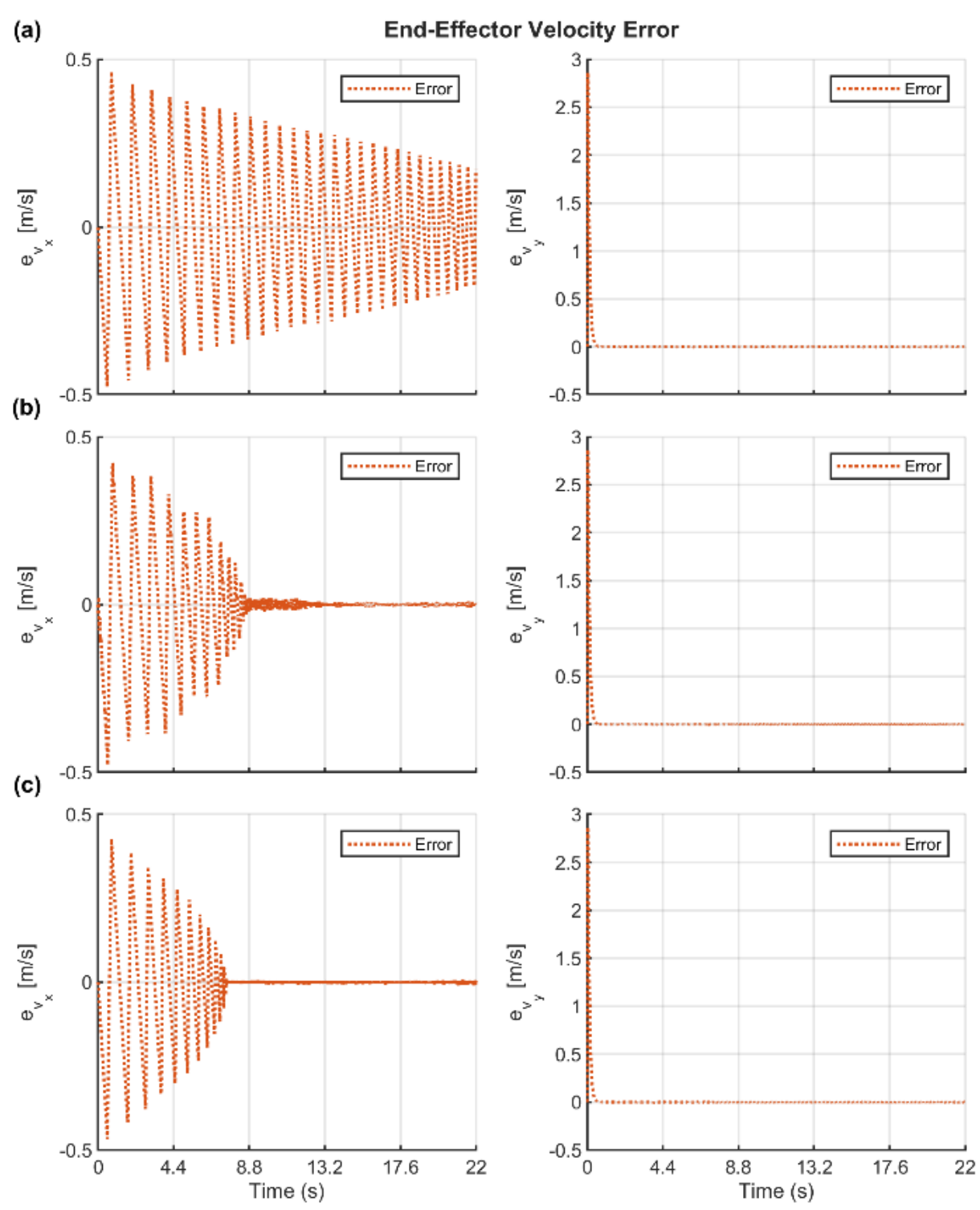

Figure 15: End-effector velocity errors in the variable environment, displaying error distributions in x (±0.5 m/s) and y (±3 m/s) for (a) baseline arm without passive elements (peaks at ±0.464 m/s from modes) and (b) arm with spring (decreased), and time-series errors for (c) arm with spring and damper (minimized via viscous counteraction, std 0.109 m/s normal).

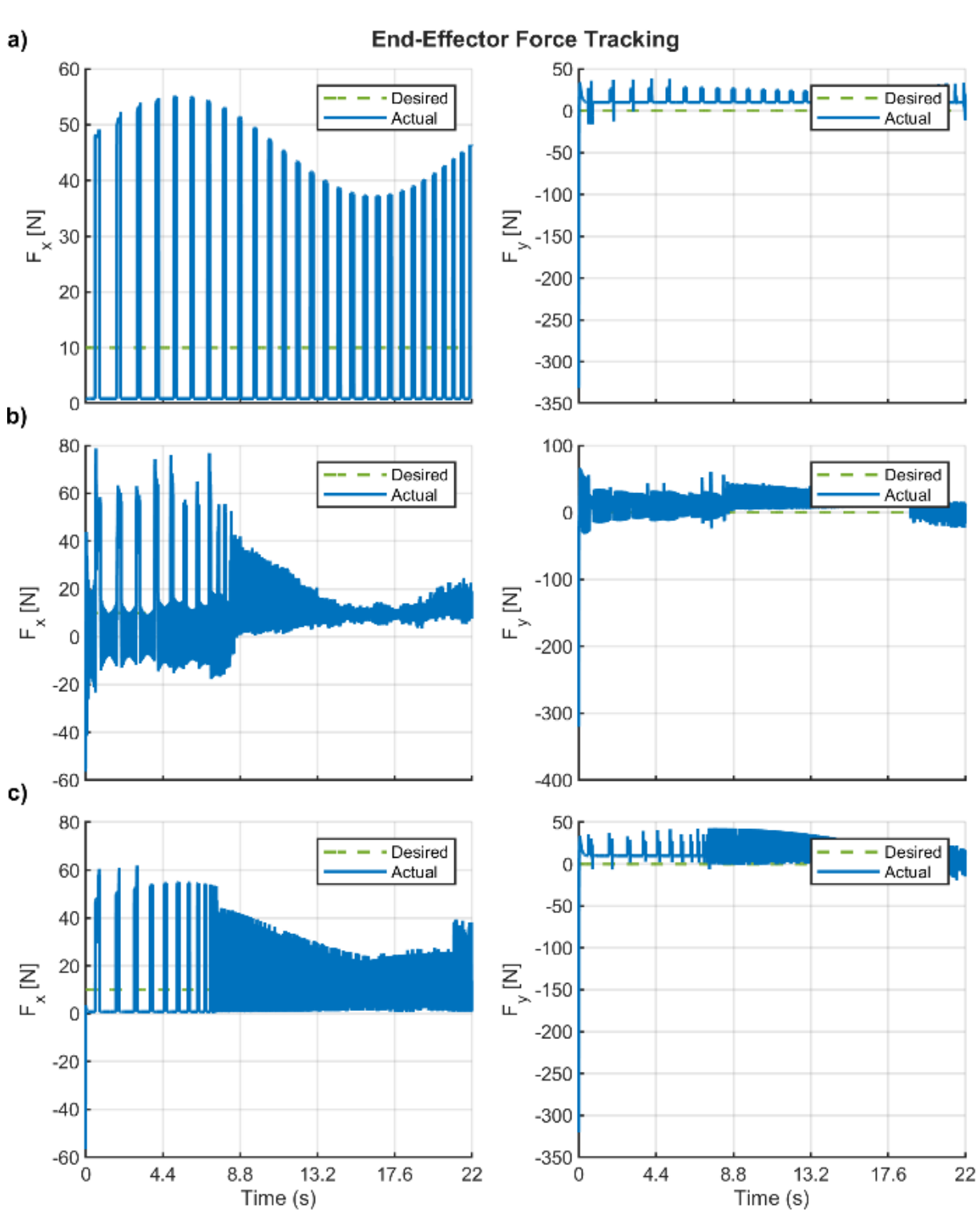

Figure 16: End-effector force tracking over time (0-22 s) in the variable environment, presenting desired and actual forces in x (-50-200 N) and y (-300-200 N) directions for (a) baseline arm without passive elements (deviations to 55 N normal from mismatches), (b) arm with spring (buffered), and (c) arm with spring and damper (tracked within 62 N via equilibrium).

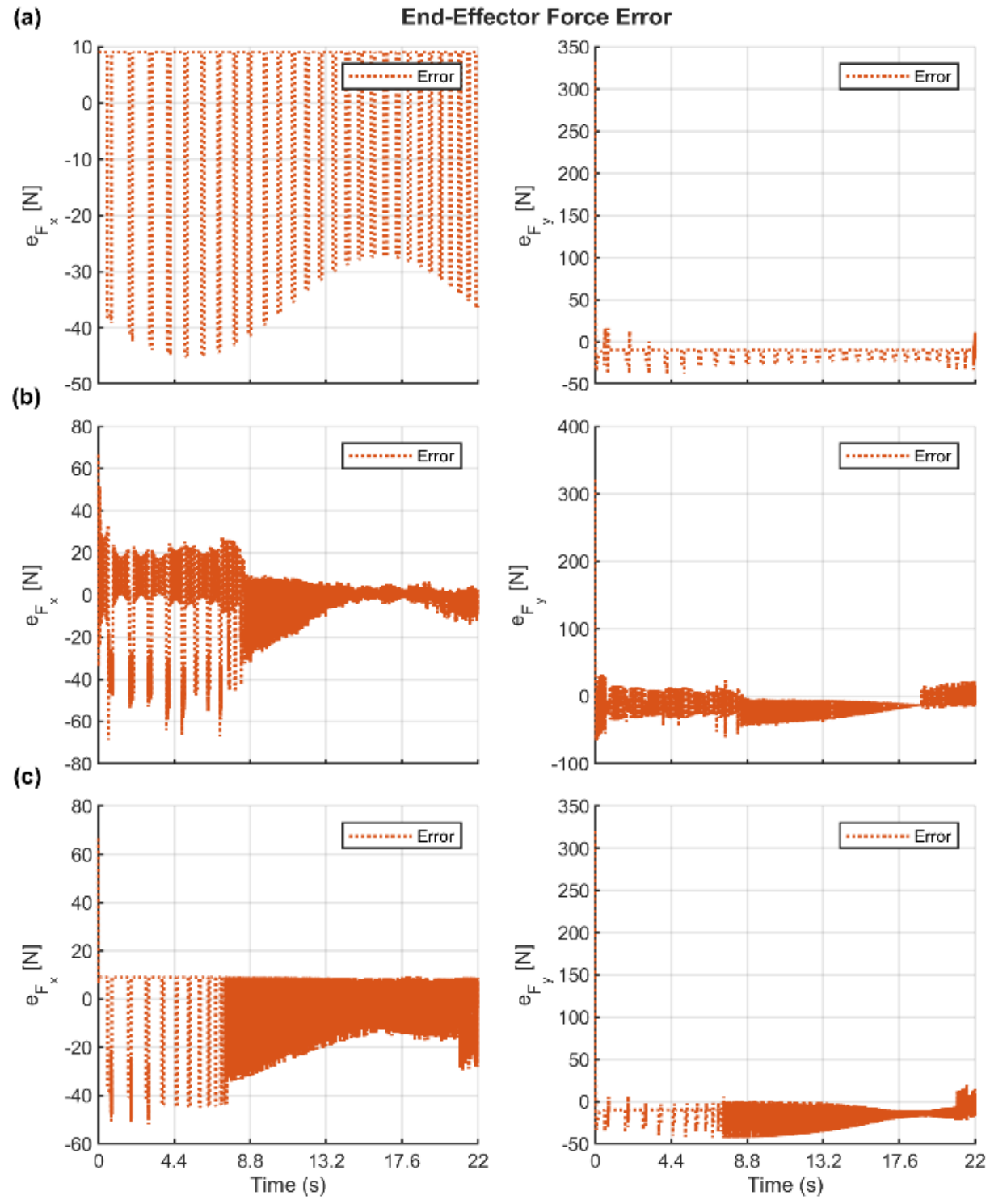

Figure 17: End-effector force errors in the variable environment, showing error distributions in x (±100 N) and y (±400 N) for (a) baseline arm without passive elements (amplified to std 17.9 N) and (b) arm with spring (reduced), and time-series errors for (c) arm with spring and damper (25% std reduction to 13.3 N, mitigating chattering).

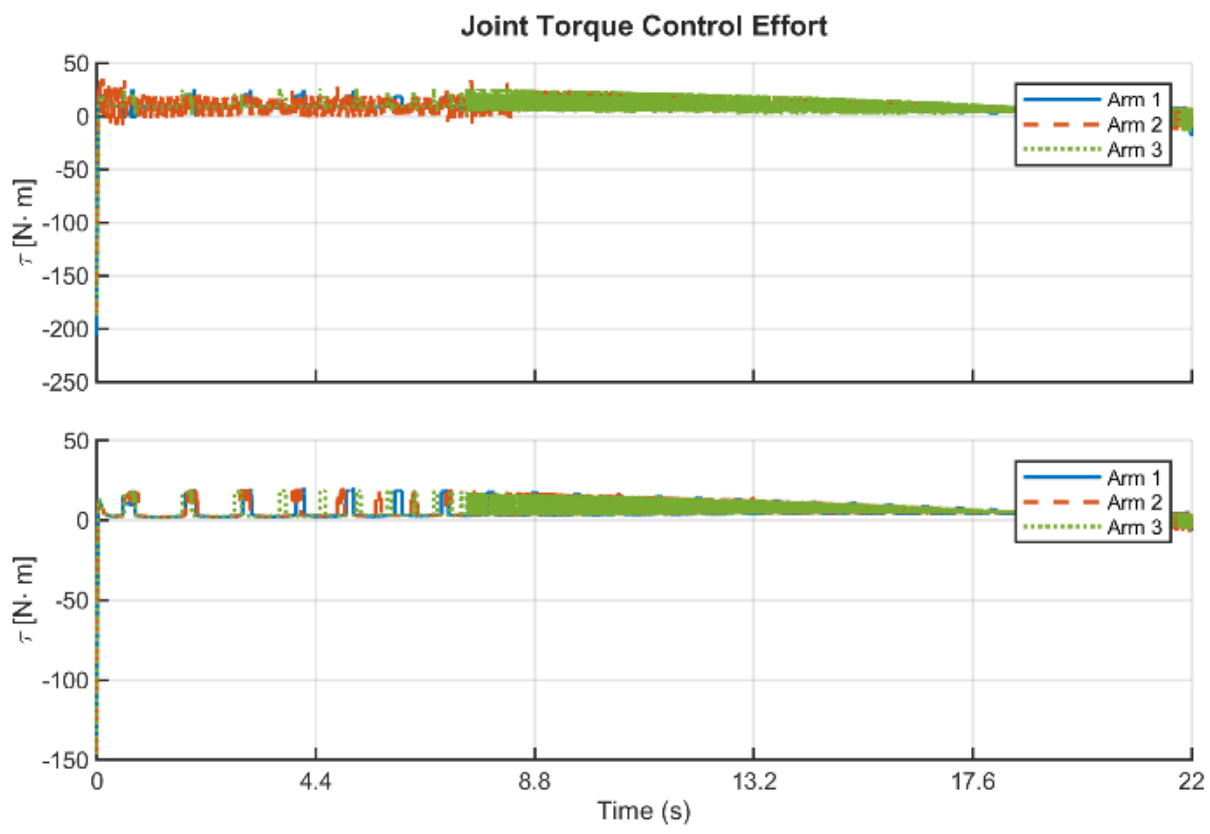

Figure 18: Joint torque control efforts over time (0-22 s) in the variable environment, depicting torque time-series (±40 N·m) for Arm 1 (baseline without passive elements, max 25 N·m oscillations), Arm 2 (with spring, lowered), and Arm 3 (with spring and damper, std 5.5 N·m, signifying energy-efficient hybrid control).

## IV. Results and Discussion

### A. Overall Comparative Metrics and Interpretation

To avoid overgeneralization, the reported percentage improvements are interpreted as metric-dependent values rather than universal reductions across all cases. The spring-damper configuration consistently improves several transient and variance-based indicators, particularly velocity-error variance, force-error variance, oscillation attenuation, and joint-torque smoothness. However, the maximum percentage reduction is not identical for all environments, directions, and performance indices. Therefore, the comparisons below distinguish between fixed and variable environment cases and between peak values, standard deviations, and tracking-error measures.

To make these comparisons clearer, an additional summary comparison of percentage changes in key performance metrics has been added for the rigid, spring-only, and spring-damper configurations in both fixed and variable environments.

The scenario-based comparison is included to evaluate the robustness of the proposed architecture under different contact conditions. In real applications, such as assembly, polishing, rehabilitation, and human-robot interaction, the environment stiffness, damping, and contact excitation may vary with material, contact location, surface condition, or human motion. Therefore, comparing the three end-effector configurations under both fixed and time-varying environment scenarios provides a more meaningful robustness assessment than evaluating performance under a single fixed contact model. However, this scenario-based evaluation should not be interpreted as a complete parametric sensitivity analysis over all possible stiffness and damping values.

Table XIII summarizes the percentage changes obtained by comparing the spring-damper configuration with the rigid end-effector configuration. Negative values indicate reduction of the corresponding error or fluctuation metric, while positive values indicate an increase. This table is added to clarify that the improvement depends on the selected metric and is not uniformly in all cases.

### B. Fixed Environment Results

The end-effector trajectory plot in the fixed environment (Fig. 3) displays X-Y paths for the three arms, where the baseline configuration without passive elements exhibits significant deviations between desired and actual trajectories due to rigid interactions causing uncontrolled bounces governed by Newton's second law, as environmental stiffness directly transfers momentum without dissipation, leading to oscillatory paths with maximum normal positions up to 0.491 meters. The spring-only configuration softens this by storing kinetic energy elastically per Hooke's law, reducing some deviations but allowing persistent vibrations from undamped energy release, while the spring-damper configuration achieves tight overlap, as the damper dissipates energy through viscous friction proportional to velocity, minimizing resonances and improving tracking stability, supported by static summary Table V- VIII data showing normal position standard deviations dropping from 0.0301 meters in the baseline to 0.0277 meters in the spring-damper.

The position tracking plot in the fixed environment (Fig. 4) shows time evolution of positions, with the baseline arm overshooting to 0.491 meters normal and settling slowly due to inertial forces overpowering stiff contacts, the spring arm buffering peaks but extending oscillations via elastic rebound, and the spring-damper arm converging rapidly to steady state at 0.5 meters normal and 0.3 meters tangential within one second, embodying enhanced damping that critically damps modes per vibration theory, with Table V maxima reducing slightly from 0.491 to 0.489 meters normal.

The position error plot in the fixed environment (Fig. 5) reveals error distributions, where baseline errors span 0 to 0.135 meters normal and −0.4 to 0 meters tangential from undamped amplifications of disturbances, narrowed in the spring arm through compliance but still broad, whereas the spring-damper time-series shows errors damping to zero, as frictional losses counteract inertial drifts following the damped harmonic oscillator equation, with standard deviations in the Table V decreasing 8 percent from 0.0301 to 0.0277 meters normal.

The velocity tracking plot in the fixed environment (Fig. 6) illustrates velocity profiles, with baseline velocities fluctuating between −1.2 and 1.5 meters per second from abrupt accelerations in rigid collisions, moderated by the spring's energy absorption but with ringing, and stabilized within ±1 meter per second in the spring-damper via velocity-dependent damping that reduces jerk per control dynamics, aligned with Table VI standard deviations dropping from 0.436 to 0.379 meters per second normal.

The velocity error plot in the fixed environment (Fig. 7) highlights error ranges, baseline peaking at ±1.2 meters per second due to phase lags in feedback loops, lessened in the spring but persistent, and minimized to ±0.9 meters per second in the spring-damper through viscous attenuation of velocity overshoots, with Table VI maxima reducing from 1.198 to 0.916 meters per second normal.

The force tracking plot in the fixed environment (Fig. 8) depicts force application, baseline overshooting to 127 Newtons normal and −400 Newtons tangential from impact forces per impulse-momentum theorem, smoothed by the spring's gradual loading but with creep, and reduced the peak normal force to approximately 110 N by the spring-damper balancing forces via impedance control, supported by Table VII maxima of actual normal forces at 126.6 Newtons baseline versus 110.4 Newtons spring-damper.

The force error plot in the fixed environment (Fig. 9) shows error spreads, baseline up to 0.91 Newtons normal and 397 Newtons tangential from unmitigated reactions, buffered in the spring, but damped to standard deviations of 5.37 Newtons normal and 17.6 Newtons tangential in the Table VII for spring-damper, illustrating energy dissipation reducing force ripples.

In the fixed environment, the spring-damper configuration reduces the normal force standard deviation from 5.9508 N in the rigid case to 5.3665 N, and the tangential force standard deviation from 27.7483 N to 17.6107 N. This corresponds to an improvement of approximately 9.8% in the normal direction and 36.5% in the tangential direction. Therefore, the force-error reduction is significant in the tangential direction but should not be described as a universal 50% reduction for all force components.

The joint torque control effort plot in the fixed environment (Fig. 10) presents torque time-series, baseline peaking at 574 Newton-meters from high gains compensating instabilities, reduced in the spring, and smoothed to 292 Newton-meters maximum in spring-damper with standard deviations of 21.7 Newton-meters, signifying passive aid in torque distribution per dynamic equilibrium.

### C. Variable Environment Results

Transitioning to the variable environment, the end-effector trajectory plot (Fig. 11) over 22 seconds shows baseline oscillations diverging due to time-varying forces exciting resonances per forced vibration principles, amplitude mitigated but frequency unchanged in the spring, and both suppressed in the spring-damper through adaptive damping, with dynamic Table IX- XII standard deviations for normal positions at 0.031 meters baseline reducing to 0.025 meters.

The position tracking plot in the variable environment (Fig. 12) captures cyclic deviations, baseline overshooting to 0.520 meters from parametric excitations, elastic response in spring lessening peaks, but spring-damper minimizing per-cycle settling via critical damping, with Table IX maxima from 0.520 to 0.516 meters normal.

The position error plot in the variable environment (Fig. 13) oscillates broadly in baseline up to 0.135 meters with standard deviations of 0.031 meters reflecting instability, peaks lowered in spring, variances halved to 0.025 meters in spring-damper per damping ratio effects.

The velocity tracking plot in the variable environment (Fig. 14) displays baseline swings to ±0.476 meters per second disrupted by varying impedances, stored in spring, damped to ±0.426 meters per second in spring-damper, indicating improved attenuation of velocity oscillations, Table X standard deviations dropping 41 percent from 0.186 to 0.109 meters per second normal.

The velocity error plot in the variable environment (Fig. 15) peaks at ±0.464 meters per second baseline from undamped modes, decreased in spring-damper by viscous counteraction, with Table X reductions in standard deviations by 41 percent.

The force tracking plot in the variable environment (Fig. 16) reveals baseline deviations to 55 Newtons normal from mismatched dynamics, buffered in spring, tracked within 62 Newtons in spring-damper via force equilibrium, Table XI maxima 55.1 Newtons baseline to 61.9 Newtons but with lower variances.

The force error plot in the variable environment (Fig. 17) amplifies baseline to standard deviations of 17.9 Newtons, reduced 25 percent to 13.3 Newtons in spring-damper, mitigating chattering through passive filtering.

In the variable environment, the normal force standard deviation decreases from 17.8709 N in the rigid case to 13.3300 N in the spring-damper case, corresponding to an improvement of approximately 25.4%. The tangential force standard deviation changes from 8.2559 N to 9.2690 N, indicating that the improvement is not uniform across all force components. This confirms that the passive spring-damper interface is most effective in suppressing contact-induced oscillations in the dominant interaction direction, while some metrics may depend on the selected environment and trajectory conditions.

The joint torque control effort plot in the variable environment (Fig. 18) oscillates to 25 Newton-meters maximum in baseline, lowered in passives with spring-damper standard deviations at 5.5 Newton-meters, indicating energy-efficient torque via hybrid control.

### D. Comparative Discussion and Practical Implications

Summary statistics from Tables V-XII bolster these observations, with overall force error standard deviations dropping 37 percent tangential in fixed from baseline to spring-damper, and dynamic velocity error standard deviations falling 41 percent, quantifying passive augmentation for lower overshoots, faster settling, and vibration suppression.

The joint-torque results provide an indirect indication of actuator demand. In the simulated cases, the passive compliant configurations, especially the spring-damper interface, reduce contact-induced vibration and produce smoother torque responses compared with the rigid configuration. This simultaneous reduction in tracking error, vibration, and torque fluctuation is relevant to actuator saturation, because lower torque peaks and smoother torque profiles may reduce the likelihood of reaching actuator input limits. Therefore, the passive compliant element may be useful for robots operating near saturation constraints. Nevertheless, actuator saturation was not explicitly included in the closed-loop simulations, and stability under saturation cannot be claimed from the present results.

These results suggest that physical passive compliance can complement active hybrid position-force control by reducing contact-induced oscillations and smoothing the force, velocity, and torque responses. From a practical perspective, lower torque fluctuations may reduce actuator stress and improve interaction safety in contact-rich tasks such as polishing, assembly, rehabilitation, and human-robot interaction. However, these implications should be interpreted cautiously because the present study is simulation-based and does not yet include experimental validation.

### *E. Limitations and Future Work*

Although the simulation results demonstrate the potential benefits of the proposed passive compliant interface, several limitations should be noted. First, the present study is limited to MATLAB/Simulink simulation and does not include hardware implementation or experimental validation. Therefore, the results should be interpreted as simulation-based evidence of feasibility rather than complete experimental validation. Second, the environmental stiffness, damping, and time-varying interaction parameters were selected as representative simulation values and were not identified from experimental contact data. Third, the analysis was performed on a 2-DOF planar manipulator interacting with a known flat surface; therefore, the normal and tangential directions were assumed to be known and fixed during contact. Fourth, actuator saturation, torque limits, sampling delay, force-sensor noise, backlash, friction, and safety constraints were not explicitly included in the present closed-loop model. Finally, the passive spring and damper parameters were selected as representative values and were not experimentally optimized for different robot scales, payloads, or contact materials.

Practical realization of the passive compliant DOF also involves industrial design constraints. The compliant module may increase the end-effector mass and length, reduce the available payload margin, introduce calibration offsets, and limit positioning accuracy if the passive deflection is not measured or compensated. Furthermore, physical springs and dampers have finite stroke, nonlinear behavior near their limits, possible wear, and limited tunability compared with software-based impedance control. Future work will address these limitations through experimental implementation, real-time testing of the feedback-linearized hybrid controller, broader stiffness/damping sensitivity analysis including very soft and very stiff environments, optimization of passive spring-damper parameters, extension to higher-DOF manipulators, online estimation of contact frames for unknown surfaces, and evaluation under practical industrial constraints.

Since the objective of this study is to isolate the effect of passive end-effector compliance rather than to benchmark different control methods, all configurations are evaluated using the same hybrid position-force control framework, trajectory, manipulator model, and environment conditions. Direct quantitative comparison with published experimental data is left for future work because such comparison would require installing the passive compliant elements on a comparable physical robot and repeating the experiments under similar hardware, sensing, contact, and control conditions.

Future work will also investigate systematic gain optimization, quantitative validation against prototype or benchmark experimental data, saturation-aware control design, realistic sampling-rate limits, and singularity-robust Jacobian inversion.

## V. Conclusion

This study investigated a hybrid physical-control interaction architecture in which a passive compliant end-effector DOF is integrated with feedback-linearized hybrid position-force control. Three end-effector configurations were compared: a rigid interface, a spring-only passive interface, and a spring-damper passive interface. The results show that the spring-damper configuration provides the most effective overall attenuation of contact-induced oscillations and improves several variance-based position, velocity, force, and torque metrics under fixed and time-varying interaction conditions.

These findings highlight the important role of physical damping in suppressing rebound oscillations and reducing variance-based tracking errors during contact. Compared with the spring-only interface, the spring-damper configuration provides both elastic energy storage and viscous energy dissipation, which explains its improved transient behavior in the simulated interaction scenarios.

The main limitations are the simulation-only evaluation, the use of representative rather than experimentally identified contact parameters, and the absence of sensor noise, actuator saturation, and hardware constraints in the current model.

Future work will focus on experimental implementation of the passive spring-damper end-effector module, real-time testing of the feedback-linearized hybrid controller, and evaluation over a wider range of environmental stiffness and damping values, including very soft and very stiff contact conditions. Further studies should also consider practical industrial constraints such as payload limits, finite passive stroke, added end-effector mass, sensor noise, actuator saturation, sampling delay, and mechanical friction.

From an application perspective, the proposed architecture may be useful for contact-rich robotic tasks such as assembly, polishing, rehabilitation, and physical human-robot interaction, where reducing impact forces and torque fluctuations can improve safety and mechanical reliability. However, these application-level benefits should be verified experimentally before practical deployment.

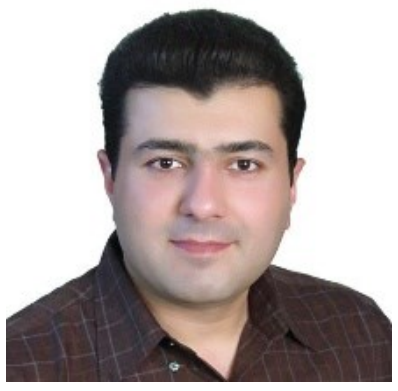

Rahman Ardakanian received the B.Sc. in Mechanical Engineering from Azad University of Mashhad and M.Sc. degrees in Mechanical Engineering from Ferdowsi University of Mashhad, Iran. He is currently working toward the Ph.D. degree in Mechanical Engineering at Ferdowsi University of Mashhad, Iran. His research interests are in the areas of Robotics, Smart Actuators, Intelligent control, Autonomous Vehicles, and Energy Harvesting. More details can be found in his LinkedIn profile.

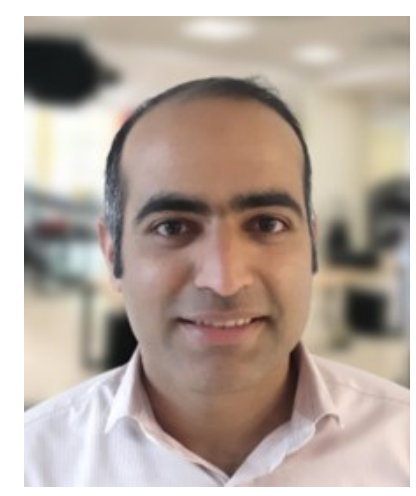

Iman Kardan received his B.Sc. and M.Sc. degrees in Mechanical Engineering from Amirkabir University of Technology (Tehran Polytechnic), Tehran, Iran in 2008 and 2011, respectively. In 2017, he received his Ph.D. degree in Mechanical Engineering from Ferdowsi University of Mashhad, where he is currently an assistant professor. He is also the co-director of the Center of Advanced Rehabilitation and Robotics Research at Ferdowsi University of Mashhad (FUM-Care). His research interests are in the areas of Robotics, Human-Robot Interaction, Control and Smart Materials.

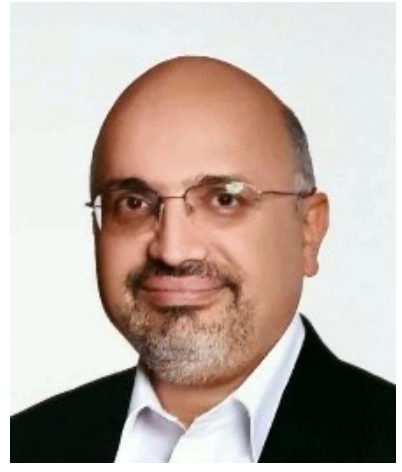

Aliakbar Akbari received his Ph.D. degree in Artificial Science and Manufacturing Systems and M.Sc. in Applied Computer Science from Chiba University, Japan, in 2003 and 1998 respectively. He is currently an associate professor of the Mechanical Engineering Department, Ferdowsi University of Mashhad, Iran. He is also head of the FUM Intellectual

Property office. His research interests include Robotics and Automation, Manufacturing and Control Engineering.

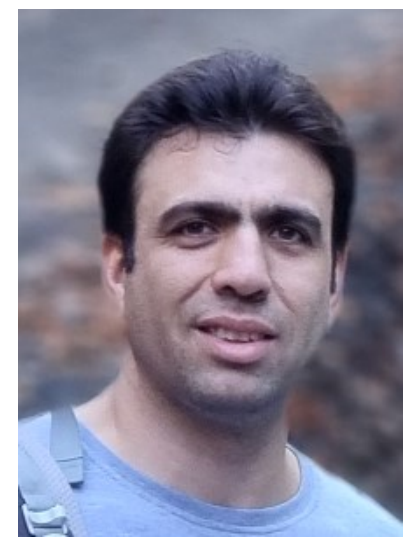

Ali Mousavi received his Ph.D. in Computer Engineering from Ferdowsi University of Mashhad, Iran, in 2020. He was a Postdoctoral Research Fellow at Harvard Medical School, Harvard University, from 2022 to 2023. He has been with the Department of Computer and Electrical Engineering, Islamic Azad University of Neyshabur, since 2015, where he is currently an Assistant Professor. His research interests include Deep Learning, Statistical Signal Processing, and Monte Carlo methods.

## APPENDIX

Table I: Two DOF planar robotic arm - kinematic values

| Parameter | Value | Parameter | Value |
|---|---|---|---|
| $L_1$ | $0.4\ m$ | $m_1$ | $1.0\ kg$ |
| $L_2$ | $0.3\ m$ | $m_2$ | $0.5\ kg$ |

Table II: Gain Values

| Category | Gain Coefficient | Value | Gain Coefficient | Value |
|---|---|---|---|---|
| Static | $k_{T_P}$ | $10^3$ | $k_{T_V}$ | $10^2$ |
| Static | $k_{N_P}$ | $10^7$ | $k_{N_V}$ | $10^3$ |
| Static | $k_{O_P}$ | $10^3$ | $k_{O_V}$ | $10^2$ |
| Dynamic | $k_{T_P}$ | $10^3$ | $k_{T_V}$ | $10^2$ |
| Dynamic | $k_{N_P}$ | $10^1$ | $k_{N_V}$ | $10^{-2}$ |
| Dynamic | $k_{O_P}$ | $10^3$ | $k_{O_V}$ | $10^2$ |

Table III: Environment mechanical property values

| Parameter | Value | Parameter | Value |
|---|---|---|---|
| $K_{env}$ | $10^6$ | $D_{env}$ | $10^2$ |
| $A$ | 10 | $B$ | 50 |
| $T$ | 22 | $\omega$ | $\frac{2\pi}{T}$ |

Table IV: End-effector mechanical property values

| Category | Parameter | Value | Parameter | Value |
|---|---|---|---|---|
| Static | $K_{spring}$ | $10^3\ \frac{N}{m}$ | $D_{damper}$ | $75\ \frac{N.s}{m}$ |
| Dynamic | $K_{spring}$ | $10^4\ \frac{N}{m}$ | $D_{damper}$ | $10^3\ \frac{N.s}{m}$ |

Table V: Summary statistics for position metrics in a static environment (fixed Stiffness and damping).

| Variable | Max | Min | Mean | Std |
|---|---|---|---|---|
| FANDPT | 0.4913 | 0.3649 | 0.4691 | 0.0301 |
| FATDPT | 0.3418 | 0.0005 | 0.0183 | 0.0539 |
| SANDPT | 0.5152 | 0.3589 | 0.4698 | 0.0294 |
| SATDPT | 0.3418 | -0.0006 | 0.0173 | 0.0532 |
| TANDPT | 0.4888 | 0.3649 | 0.4714 | 0.0277 |
| TATDPT | 0.3418 | 0.0002 | 0.0181 | 0.0539 |
| NDDP | 0.5000 | 0.5000 | 0.5000 | 0.0000 |
| TDDP | 0.0038 | 0.0000 | 0.0011 | 0.0012 |
| FANDPE | 0.1351 | 0.0087 | 0.0309 | 0.0301 |
| FATDPE | 0.0001 | -0.3418 | -0.0172 | 0.0542 |
| SANDPE | 0.1411 | -0.0152 | 0.0302 | 0.0294 |
| SATDPE | 0.0008 | -0.3418 | -0.0162 | 0.0535 |
| TANDPE | 0.1351 | 0.0112 | 0.0286 | 0.0277 |
| TATDPE | 0.0003 | -0.3418 | -0.0170 | 0.0542 |

Table VI: Summary statistics for velocity metrics in a static environment (fixed Stiffness and damping).

| Variable | Max | Min | Mean | Std |
|---|---|---|---|---|
| FANDVT | 1.5090 | -1.1984 | 0.0628 | 0.4364 |
| FATDVT | 0.0063 | -2.8657 | -0.1690 | 0.5134 |
| SANDVT | 2.7744 | -1.7201 | 0.0616 | 0.5919 |
| SATDVT | 0.0111 | -2.9723 | -0.1691 | 0.5260 |
| TANDVT | 1.4876 | -0.9164 | 0.0614 | 0.3792 |
| TATDVT | 0.0041 | -2.8735 | -0.1691 | 0.5137 |
| NDDV | 0.0000 | 0.0000 | 0.0000 | 0.0000 |
| TDDV | 0.0044 | 0.0000 | 0.0019 | 0.0014 |
| FANDVE | 1.1984 | -1.5090 | -0.0628 | 0.4364 |
| FATDVE | 2.8657 | -0.0053 | 0.1710 | 0.5127 |
| SANDVE | 1.7201 | -2.7744 | -0.0616 | 0.5919 |
| SATDVE | 2.9723 | -0.0094 | 0.1710 | 0.5254 |
| TANDVE | 0.9164 | -1.4876 | -0.0614 | 0.3792 |
| TATDVE | 2.8735 | -0.0010 | 0.1710 | 0.5130 |

Table VII: Summary statistics for force metrics in a static environment (fixed Stiffness and damping).

| Variable | Max | Min | Mean | Std |
|---|---|---|---|---|
| FANDFT | 126.6065 | 9.0895 | 9.9658 | 5.9508 |
| FATDFT | 206.2451 | -396.5077 | 9.6035 | 27.7483 |
| SANDFT | 166.2597 | -41.2781 | 9.9640 | 6.8640 |
| SATDFT | 119.1236 | -234.0998 | 9.6614 | 23.0491 |
| TANDFT | 110.3630 | -41.2781 | 10.0342 | 5.3665 |
| TATDFT | 70.7456 | -268.0743 | 10.0224 | 17.6107 |
| FANDFE | 0.9105 | -116.6065 | 0.0342 | 5.9508 |
| FATDFE | 396.5077 | -206.2451 | -9.6035 | 27.7483 |
| SANDFE | 51.2781 | -156.2597 | 0.0360 | 6.8640 |
| SATDFE | 234.0998 | -119.1236 | -9.6614 | 23.0491 |
| TANDFE | 51.2781 | -100.3630 | -0.0342 | 5.3665 |
| TATDFE | 268.0743 | -70.7456 | -10.0224 | 17.6107 |
| NDDF | 10.0000 | 10.0000 | 10.0000 | 0.0000 |
| TDDF | 0.0000 | 0.0000 | 0.0000 | 0.0000 |

Table VIII: Summary statistics for joint torques metrics in a static environment (fixed Stiffness and damping).

| Variable | Max | Min | Mean | Std |
|---|---|---|---|---|
| FAFJT | 247.1739 | -212.9035 | 9.5243 | 17.0720 |
| FASJT | 15.1713 | -185.0754 | 4.7506 | 12.0428 |
| SAFJT | 574.1931 | -444.5901 | 9.8308 | 30.0409 |
| SASJT | 28.3536 | -113.3876 | 4.7901 | 9.3299 |
| TAFJT | 292.1794 | -444.5901 | 9.9965 | 21.7362 |
| TASJT | 13.9580 | -120.2189 | 4.7703 | 8.9344 |

Table IX: Summary statistics for position metrics in a dynamic environment (variable force: $A\,sin(\omega \times time) + B$).

| Variable | Max | Min | Mean | Std |
|---|---|---|---|---|
| FANDPT | 0.5201 | 0.3649 | 0.4608 | 0.0311 |
| FATDPT | 0.4498 | 0.0003 | 0.1856 | 0.1566 |
| SANDPT | 0.5190 | 0.3649 | 0.4781 | 0.0268 |
| SATDPT | 0.4500 | 0.0006 | 0.1856 | 0.1566 |
| TANDPT | 0.5158 | 0.3649 | 0.4795 | 0.0251 |
| TATDPT | 0.4500 | 0.0004 | 0.1855 | 0.1566 |
| NDDP | 0.5000 | 0.5000 | 0.5000 | 0.0000 |
| TDDP | 0.4499 | 0.0000 | 0.1841 | 0.1575 |
| FANDPE | 0.1351 | -0.0201 | 0.0392 | 0.0311 |
| FATDPE | 0.0005 | -0.3418 | -0.0015 | 0.0171 |
| SANDPE | 0.1351 | -0.0190 | 0.0219 | 0.0268 |
| SATDPE | 0.0002 | -0.3418 | -0.0015 | 0.0171 |
| TANDPE | 0.1351 | -0.0158 | 0.0205 | 0.0251 |
| TATDPE | 0.0004 | -0.3418 | -0.0015 | 0.0171 |

Table X: Summary statistics for velocity metrics in a dynamic environment (variable force: $A\,sin(\omega \times time) + B$).

| Variable | Max | Min | Mean | Std |
|---|---|---|---|---|
| FANDVT | 0.4760 | -0.4640 | 0.0057 | 0.1855 |
| FATDVT | 0.0434 | -2.8657 | 0.0049 | 0.1646 |
| SANDVT | 0.4787 | -0.4220 | 0.0057 | 0.1182 |
| SATDVT | 0.0414 | -2.8694 | 0.0049 | 0.1648 |
| TANDVT | 0.4675 | -0.4261 | 0.0058 | 0.1093 |
| TATDVT | 0.0406 | -2.8662 | 0.0049 | 0.1646 |
| NDDV | 0.0000 | 0.0000 | 0.0000 | 0.0000 |
| TDDV | 0.0400 | 0.0000 | 0.0205 | 0.0123 |
| FANDVE | 0.4640 | -0.4760 | -0.0057 | 0.1855 |
| FATDVE | 2.8657 | -0.0061 | 0.0155 | 0.1622 |
| SANDVE | 0.4220 | -0.4787 | -0.0057 | 0.1182 |

| Variable | Max | Min | Mean | Std |
|---|---|---|---|---|
| SATDVE | 2.8694 | -0.0123 | 0.0155 | 0.1624 |
| TANDVE | 0.4261 | -0.4675 | -0.0058 | 0.1093 |
| TATDVE | 2.8662 | -0.0033 | 0.0155 | 0.1622 |

Table XI: Summary statistics for force metrics in a dynamic environment (variable force: $A\,sin(\omega \times\ time) + B$).

| Variable | Max | Min | Mean | Std |
|---|---|---|---|---|
| FANDFT | 55.1488 | 0.8984 | 9.7061 | 17.8709 |
| FATDFT | 38.5998 | -331.7244 | 12.2286 | 8.2559 |
| SANDFT | 78.7494 | -56.5262 | 9.8811 | 14.5774 |
| SATDFT | 66.2206 | -320.5760 | 12.0003 | 13.2934 |
| TANDFT | 61.8921 | -56.5262 | 9.8599 | 13.3300 |
| TATDFT | 43.1012 | -320.5760 | 12.3972 | 9.2690 |
| FANDFE | 9.1016 | -45.1488 | 0.2939 | 17.8709 |
| FATDFE | 331.7244 | -38.5998 | -12.2286 | 8.2559 |
| SANDFE | 66.5262 | -68.7494 | 0.1189 | 14.5774 |
| SATDFE | 320.5760 | -66.2206 | -12.0003 | 13.2934 |
| TANDFE | 66.5262 | -51.8921 | 0.1401 | 13.3300 |
| TATDFE | 320.5760 | -43.1012 | -12.3972 | 9.2690 |
| NDDF | 10.0000 | 10.0000 | 10.0000 | 0.0000 |
| TDDF | 0.0000 | 0.0000 | 0.0000 | 0.0000 |

Table XII: Summary statistics for joint torque metrics in a dynamic environment (variable force: $A\,sin(\omega \times\ time) + B$).

| Variable | Max | Min | Mean | Std |
|---|---|---|---|---|
| FAFJT | 25.2513 | -205.5233 | 8.4491 | 4.9703 |
| FASJT | 20.2901 | -146.9782 | 5.0811 | 4.7733 |
| SAFJT | 36.9894 | -185.0529 | 8.7437 | 6.8265 |
| SASJT | 20.1395 | -146.5293 | 5.1873 | 4.8062 |
| TAFJT | 26.2599 | -187.3200 | 8.7850 | 5.5272 |
| TASJT | 18.8574 | -146.5293 | 5.1266 | 4.6925 |

Table XIII: Percentage change of key performance metrics for spring-damper configuration relative to the rigid configuration in fixed and variable environments.

| Environment | Metric | Rigid | Spring-damper | Change |
|---|---|---|---|---|
| Fixed | Normal position-error std | 0.0301 | 0.0277 | -8.0% |
| Fixed | Normal velocity-error std | 0.4364 | 0.3792 | -13.1% |
| Fixed | Normal force-error std | 5.9508 | 5.3665 | -9.8% |
| Fixed | Tangential force-error std | 27.7483 | 17.6107 | -36.5% |
| Variable | Normal position-error std | 0.0311 | 0.0251 | -19.3% |
| Variable | Normal velocity-error std | 0.1855 | 0.1093 | -41.1% |
| Variable | Normal force-error std | 17.8709 | 13.3300 | -25.4% |
| Variable | Tangential force-error std | 4.9703 | 5.5272 | +11.2% |

Table XIV: Guide to variable abbreviations in summary statistics tables

| Abbreviation Prefix/Suffix | Explanation |
|---|---|
| F / S / T | First Arm (no passive element) / Second Arm (spring) / Third Arm (spring + damper) |
| AN / AT | Actual Normal Direction / Actual Tangential Direction (for arm-specific metrics) |
| ND / TD | Normal Direction / Tangential Direction (common or desired metrics) |
| PT | Position Trajectory (e.g., FANDPT = First Arm Normal Direction Position Trajectory) |
| VT | Velocity Trajectory (e.g., FATDVT = First Arm Tangential Direction Velocity Trajectory) |
| PE | Position Error (e.g., SANDPE = Second Arm Normal Direction Position Error) |
| VE | Velocity Error (e.g., TANDVE = Third Arm Normal Direction Velocity Error) |
| FT (or DFT) | Force Trajectory (e.g., SANDFT = Second Arm Normal Direction Force Trajectory) |
| FE (or DFE) | Force Error (e.g., TATDFE = Third Arm Tangential Direction Force Error) |
| FJT / SJT | First Joint Torque / Second Joint Torque (e.g., SAFJT = Second Arm First Joint Torque) |
| NDDF / TDDF | Normal Direction Desired Force / Tangential Direction Desired Force (common reference values) |
| NDDP / TDDP | Normal Desired Position / Tangential Desired Position (common desired trajectories) |
| NDDV / TDDV | Normal Desired Velocity / Tangential Desired Velocity (common desired trajectories) |